# A Review on Machine Learning Approaches for the Prediction of Glucose Levels and Hypogylcemia

Beyza Cinar[1*], Louisa van den Boom[2], and Maria Maleshkova[1]

[1]*Department of Data Engineering, Helmut-Schmidt-University, Hamburg, Germany,*
[2]*Helios Klinikum Gifhorn GmbH, Clinic for Pediatric and Adolescent Medicine, Gifhorn, Germany*

*Corresponding author
Email address: cinarb@hsu-hh.de

**Abstract:** Type 1 Diabetes (T1D) is an autoimmune disease leading to insulin insufficiency. Thus, patients require lifelong insulin therapy, which has a side effect of hypoglycemia. Hypoglycemia is a critical state of decreased blood glucose levels (BGL) below 70 mg/dL and is associated with increased risk of mortality. Machine learning (ML) models can improve diabetes management by predicting hypoglycemia and providing optimal prevention methods. ML models are classified into regression and classification based, that forecast glucose levels and identify events based on defined labels, respectively. This review investigates state-of-the-art models trained on data of continuous glucose monitoring (CGM) devices from patients with T1D. We compare the models' performance across short-term (15 to 120 min) and long term (3 to more than 24 hours) prediction horizons (PHs). Particularly, we explore: 1) How much in advance can glucose values or a hypoglycemic event be accurately predicted? 2) Which models have the best performance? 3) Which factors impact the performance? and 4) Does personalization increase performance? The results show that 1) a PH of up to 1 hour provides the best results. 2) Conventional ML methods yield the best results for classification and DL for regression. A single model cannot adequately classify across multiple PHs. 3) The model performance is influenced by multivariate datasets and the input sequence length (ISL). 4) Personal data enhances performance but due to limited data quality population-based models are preferred.

## I. Introduction

Type 1 diabetes (T1D) is an incurable autoimmune disease and is characterized by insufficient insulin production [1–3]. Patients with T1D have increased blood glucose levels (BGL) and typically depend on insulin replacement therapies [1,4]. Critical side effects include hypoglycemia and severe hypoglycemia defined by BGL below 70 and 54 mg/dL, respectively [5]. Hypoglycemia can be life-threatening especially if unnoticed since it can lead to unconsciousness, coma, organ damage, and to death in the worst case [6–9]. Notably, with impaired glucose awareness, symptoms tend to occur with lower glucose thresholds [10]. Continuous glucose monitoring (CGM) devices, which measure glucose in the intestinal fluid, can decrease the occurrence of hypoglycemic events. Moreover, utilizing machine learning (ML) models, these devices can predict anomalies or simulate future values [11,4,12]. Prediction tasks are typically part of supervised learning methods, which are commonly classified into classification- and regression-based approaches [13,14]. A classification model analyzes the patterns in data and predicts the annotated class. In contrast, regression models forecast a sequence of numerical data [14]. While both approaches are applicable, the onset of hypoglycemia could be better predicted with classification models, since the focus is on the condition of interest. Studies report that classification approaches lead to fewer variations [15], and better recall but worse precision [16]. Contrariwise, forecasting glucose values enables the simulation of glucose sequences under different scenarios, and can support the automation of insulin pumps, and the management of diet and exercise [17,18].
Several factors influence the variations in glucose levels, complicate the regulation of adequate insulin doses and could impact the performance of prediction models [19,20]. Depending on the use case, a different prediction horizon (PH) may be required for the system, since each action causing low BGL could have a different impact duration on the glucose metabolism. For instance, a long-term PH from 3-42 h could mitigate insulin and exercise-induced hypoglycemia because bolus insulin can have a duration of 3-8 h, basal insulin of 20-42 h [4,6,21], and exercise could lead to side effects even after 24 h [22]. Moreover, a short-term PH of 30-120 min could enable preventive self-actions.

The aim of this review is to explore the current state of glucose forecasting and hypoglycemia classification. Subsequently, we focus on the following questions: 1) How much in advance can glucose values or a hypoglycemic event be accurately predicted? 2) Which models have the best performance? 3) Which factors impact the performance? and 4) Does personalization increase performance? The presented models are examined for corresponding PHs classified into short- and long-term defined as 15 to 120 min and 3 to more than 24 h, respectively.
The remainder of this work is as follows. Section II introduces the methodology. Section III presents the results and compares the studies. The findings and identified research gaps are discussed in Section IV. Finally, Section V concludes this review.

## II. Methods

This review aims to determine effective model architectures and configurations for precise hypoglycemia prediction in T1D research. The motivation is to develop a multi-functional model capable of accurately detecting short- and long-term hypoglycemic events. The objective is to enhance glucose management and minimize hypoglycemia risk. Accordingly, recent studies are compared in terms of the applied methods, datasets, and performance.
The systematic literature review followed the PRISMA [23] method presented in Fig. 1. Related works were identified in computer science and medical databases such as "ScienceDirect", "IEEE Xplore", "PubMed", "ResearchGate" and on "Google Scholar" with the following search-phrase: "type 1 diabetes" AND ("hypoglycemia" OR "glucose") AND ("forecasting" OR "prediction" OR "classification") AND ("machine learning" OR "deep learning"). The search was conducted from May 6 to 11 June 2024, and search alerts were activated to not miss relevant new studies. "PubMed" illustrated an increase of studies since 2019 as to why the search was filtered to a publication date of at least 2019. Also, some remarkable works since 2013 were chosen later from reference tracking. The search was limited to English language, and reviews were excluded. For "ScienceDirect", only research articles were included, and the subject areas were filtered to computer science, mathematics, and engineering. For IEEE Xplore, it could be filtered to T1D. After title screening, "Google Scholar" presented additional 39 studies. For "ResearchGate", the results were ordered based on relevance and after screening the first 100 entries, 8 studies were included.
Inclusion criteria were studies forecasting glucose or classifying hypoglycemia with ML or deep learning (DL) methods. Only studies using data of patients with T1D were included. The classification of diabetes onset was not included. Also, papers only relying on statistical or mathematical methods were removed. Studies not using CGM data or utilizing virtual data were excluded as well, since the results would not contribute to the research questions. In addition, studies with insufficient results, without any major contribution, which do not present numerical results, and which focus only on very short-term PHs were not included.

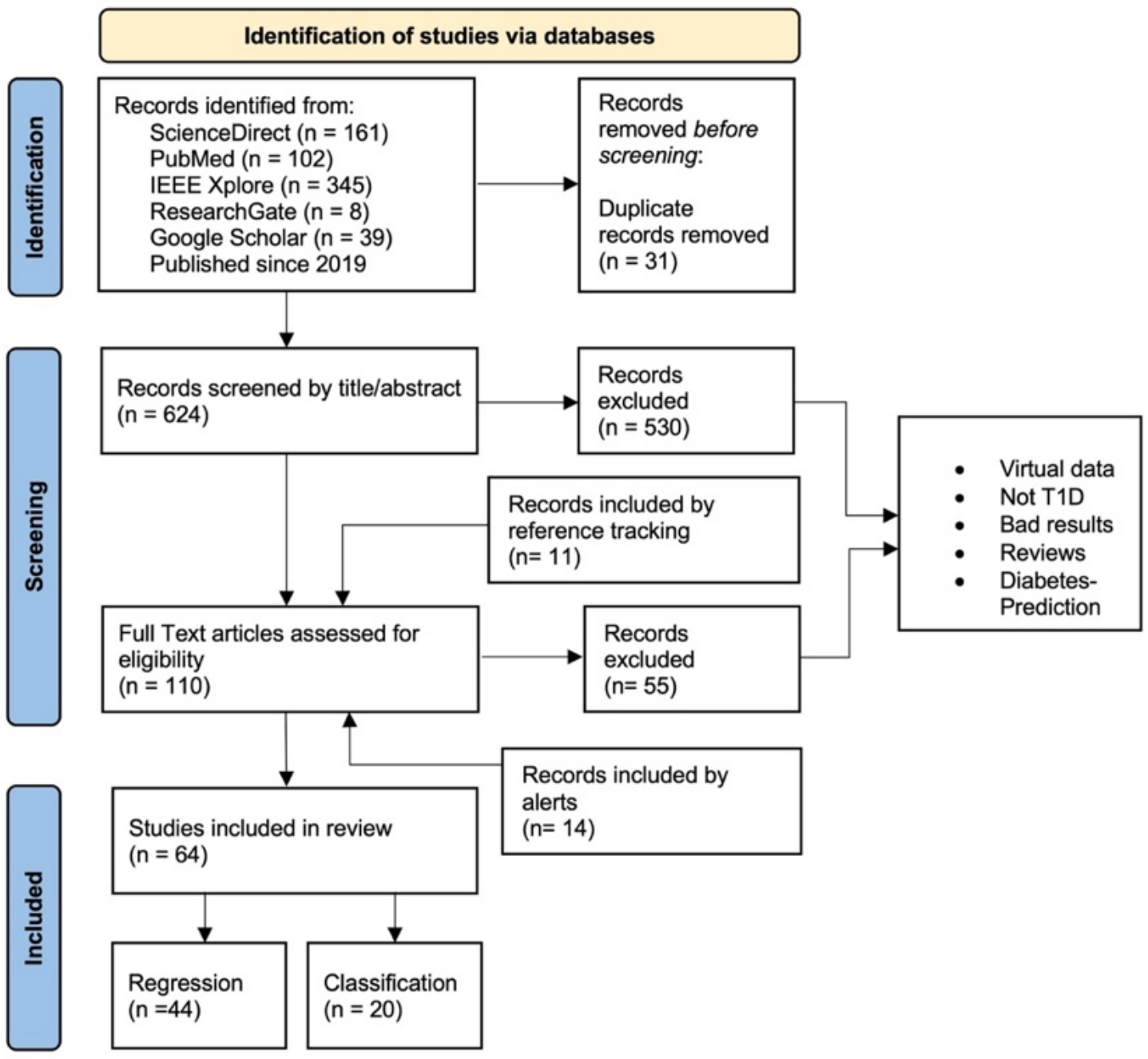

*Figure 1: Literature Selection Flow Based on the PRISMA Method*

# III. Results

This section first presents studies forecasting glucose levels (A), followed by studies classifying hypoglycemia (B). Finally, leveraged datasets are detailed (C). The focus lies on the chosen PHs and utilized ML architectures. The featured studies, input data and performance across the considered PHs are displayed in Tables 1 and 2, respectively for regression and classification models. Fig. 2 illustrates the distribution of predicted PHs across included studies. Most studies, especially regression-based approaches, focus on short-term PHs. In contrast, classification-based models more often consider longer PHs of 6 to more than 24 h before hypoglycemia.

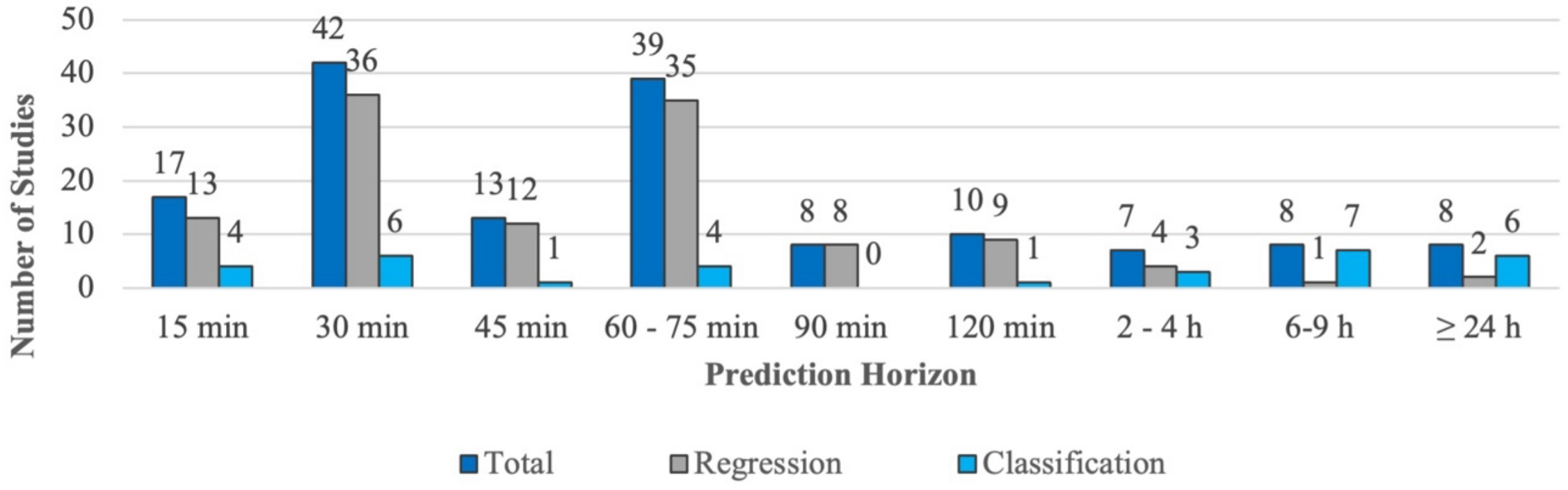

*Figure 2: Distribution of Covered Prediction Horizons. The bar chart compares the number of studies for the covered prediction horizons of regression and classification models.*

## A. Summary of Regression Models

This section reviews the latest research on glucose forecasting, focusing on PHs ranging from 15 min to 48 h. The distribution of utilized PHs, seen in Fig. 2, indicates that for regression-based models, PHs of 30 min and 60-75 min are most popular. As can be seen in Table 1, most studies evaluate multiple PHs with the same model, data, and settings. In general, studies forecast short-term glucose values. Fig. 3 illustrates the distribution of utilized ML models. Conventional ML approaches have been less frequently applied to regression tasks, accounting for only 5 studies. In contrast, DL models have become the predominant methodology and are leveraged by 23 studies. EL and more complex hybrid architectures have also gained increasing popularity, comprising 16 studies.

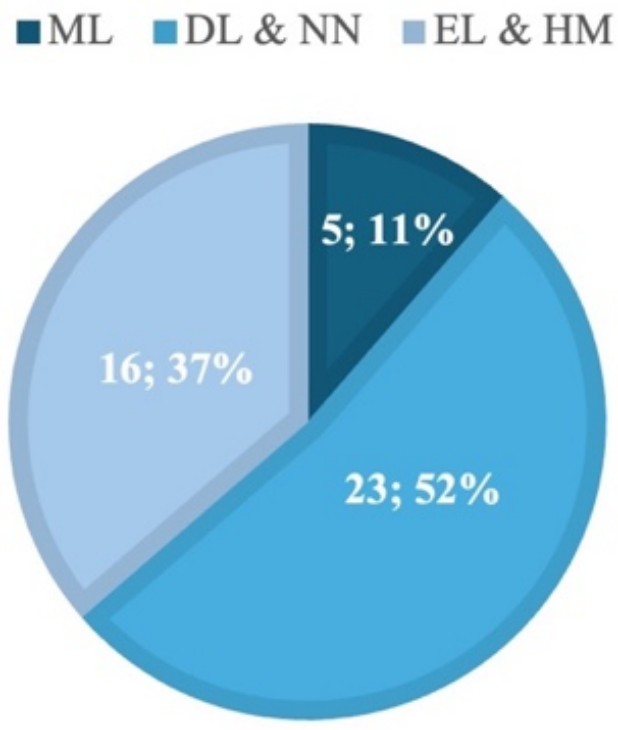

*Figure 3: Distribution of ML, DL, and EL & HM Among Regression Studies. DL, Deep Learning. EL, Ensemble Learning. HM, Hybrid Models. ML, Machine Learning. NN, Neural Networks.*

The following section compares studies that forecast glucose levels, with short-term PHs analyzed in detail to highlight the most effective methods for each PH. In addition, the stability of models that predict multiple PHs is examined. For each PH, only the most relevant and impactful studies are discussed. In contrast, long-term PHs are reviewed based on broader horizons. The studies are mainly compared based on the RMSE values if available.

*Table 1: Regression Models for Short- and Long-term PHs*

| Study | Model | Input | Features | ISL | Performance of PHs | | | | | | | | |
|---|---|---|---|---|---|---|---|---|---|---|---|---|---|
| | | | | | Short-term PH in min | | | | | | Long-term PH in h | | |
| | | | | | 15 | 30 | 45 | 60 | 90 | 120 | 3-4 | 6 -9 | ≥24 |
| Georga et al., 2013 [24] | SVR | 15 (MTB1) | GLC, insulin, CHO | 30 - 90 min | 5.21 | 6.03 | N/A | 7.14 | N/A | 7.62 | N/A | N/A | N/A |
| Zarkogianni et al., 2015 [25] | SOM | 10 (MTB2) | GLC, PA | 30 - 120 min | N/A | 11.42 | N/A | 19.58 | N/A | 31.00 | N/A | N/A | N/A |
| Ben Ali et al., 2018 [26] | FFNN | 12 (PBA) | GLC | 90-135 min | 6.43 | 7.45 | 8.13 | 9.03 | N/A | N/A | N/A | N/A | N/A |
| Martinsson et al., 2019 [27] | LSTM | 6 (O18) | GLC | 60 min | N/A | N/A | N/A | 18.87 | 31.40 | N/A | N/A | N/A | N/A |
| Alfian et al., 2020 [28] | MLP | 12 (DN07) | GLC | 30 min | 2.82 | 6.31 | 10.65 | 15.33 | N/A | N/A | N/A | N/A | N/A |
| Munoz Organero, 2020 [29] | LSTM | 40/9 (D1NAMO) | GLC, insulin, CHO | 9 h | N/A | 6.42 | N/A | 11.35 | N/A | N/A | N/A | N/A | N/A |
| Allam, 2021 [30] | FFNN + WOA | 9 | GLC | 50–100 min | 2.56 | 6.49 | 13.51 | 21.44 | N/A | N/A | N/A | N/A | N/A |
| Seo et al., 2021 [31] | CNN | 141 (SMC) | GLC, static data | 80 min | 10.9 | 17.8 | 23.3 | 28.1 | N/A | N/A | N/A | N/A | N/A |
| Daniels et al., 2022 [32] | CRNN | 12 (O18+O20) | GLC, insulin, CHO, PA | 2h | N/A | 18.80 | 25.30 | 31.80 | 41.20 | 47.20 | N/A | N/A | N/A |
| Jaloli and Cescon, 2022 [33] | CNN + LSTM | 168/59 (RGL/DA) | GLC, insulin, CHO | 3 * PH | N/A | 9.28/ 9.81 | N/A | 16.51/ 18.32 | 23.45/ 25.12 | N/A | N/A | N/A | N/A |
| Katsarou et al., 2022 [34] | LSTM | 29 (GML) | GLC | 24h? | 6.41 | 8.81 | N/A | N/A | N/A | N/A | N/A | N/A | N/A |
| Nemat et al., 2022[35] | CCMBA + LSTM | 6/6 (O18/O20) | GLC, insulin, CHO, HR, PA | 90 min | N/A | 20.09/ 19.44 | N/A | 32.90/ 34.79 | N/A | N/A | N/A | N/A | N/A |
| Phadke et al., 2022 [36] | ANN | 12 (O18+O20) | GLC, PHY and PSY data, PA, insulin, CHO | 2–6h | N/A | 0.04 | N/A | 0.07 | N/A | 0.15 | 0.21 | N/A | 0.13 |
| Tyler et al., 2022 [37] | MARS | 20 (PT) | GLC, PA, PA history | N/A | N/A | N/A | 19.7 | N/A | N/A | N/A | 23.0 | N/A | N/A |
| Zhu et al., 2022 [38] | FCNN | 12/12/25 (O18+O20/A /ABC4D) | GLC, insulin, CHO | N/A | N/A | 18.64/ 20.23/ 20.25 | N/A | 31.07/ 35.4/ 34.03 | N/A | N/A | N/A | N/A | N/A |
| Zhu et al., 2022 [39] | LR | 12 (PZ) | GLC, insulin, CHO, PHY data, PA | 60 min | 10.15 | 20.92 | 28.99 | 35.28 | N/A | N/A | N/A | N/A | N/A |
| Annuzzi et al., 2023 [40] | FFNN | 12/15 (DN4/ AI4PG) | GLC/ GLC, insulin, lipids | 30 min | 4.14/ 2.91 | 8.30/ 8.24 | 13.72/ 15.55 | 16.69/ 22.98 | N/A | N/A | N/A | N/A | N/A |
| Bi and Kar, 2023 [9] | LSTM+ RF+ XGB | 30 (DN07) | GLC | 25-855 min | N/A | N/A | N/A | 3.86 | N/A | 4.39 | 4.99 | 5.10 | N/A |
| Cai et al., 2023 [41] | GPCEM-PSO-BP | 7 (CT-100)/ 3 | GLC | 45 min | N/A | N/A | N/A | N/A | 6.46 | N/A | N/A | N/A | N/A |
| Domanski et al., 2023 [42] | Deep RL | 12 (O18+O20) | GLC | 30 min | N/A | 18.32 | 23.43 | N/A | N/A | N/A | N/A | N/A | N/A |
| Marx et al., 2023 [43] | LSTM+/ RNN+ | 13 (PM) | GLC, insulin, CHO, PA, static data | 2h | N/A | 27.03/ 27.93 | N/A | 39.28/ 37.84 | N/A | 54.23/ 48.65 | N/A | N/A | N/A |
| Nasir et al., 2023 [44] | LSTM | 12 (DN07) | GLC, time | 30 min | 2.23 | 7.38 | N/A | N/A | N/A | N/A | N/A | N/A | N/A |
| Rajagopal and Thangarasu, 2023 [45] | Autoen-coder+ LSTM | 12 (ST1D) | GLC, insulin, CHO, age | N/A | N/A | N/A | N/A | 6.30 | 9.40 | N/A | N/A | N/A | N/A |
| Ramachandran et al., 2023 [46] | CRNN | 12 (O18+20) | GLC, insulin, CHO | N/A | N/A | N/A | N/A | 4.21 | N/A | N/A | N/A | N/A | N/A |
| Shuvo and Islam, 2023 [47] | LSTM | 12 (O18+O20) | GLC, insulin, CHO | 2h | N/A | 16.06 | N/A | 30.89 | 40.51 | 47.39 | N/A | N/A | N/A |
| Yang et al., 2023 [48] | PBGTAM | 12 (O18+O20) | GLC, insulin, CHO | N/A | N/A | 20.57 | N/A | 35.09 | N/A | N/A | N/A | N/A | N/A |
| Yang et al., 2023 [49] | Ensemble of 3 LSTM models | 12 (O18+O20) | GLC | 30-90 min | N/A | 1.42 | 3.21 | 6.35 | N/A | N/A | N/A | N/A | N/A |
| Zafar et al., 2023 [50] | LSTM | 19 (OpenAPS) | GLC | 60 days | N/A | N/A | N/A | N/A | N/A | N/A | N/A | N/A | 3.7 |
| Allam, 2024 [51] | NNARX | 9 | GLC | 100 min | 21.42 | 16.43 | 13.42 | 13.53 | 18.39 | N/A | N/A | N/A | N/A |
| Floris and Vasata, 2024 [52] | LMU | 6/6 (O18/O20) | GLC, insulin, CHO | 30 min | N/A | 18.17/ 18.56 | N/A | 30.33/ 32.57 | N/A | N/A | N/A | N/A | N/A |
| Ghimire et al., 2024 [53] | LSTM | 12/112/101/ 451 (O18+O20)/ PBro/ PBre/RT | GLC | 2 h | N/A | 18.26/ 20.33/ 26.38/ 26.31 | N/A | 31.12/ 34.91/ 43.66/ 41.94 | N/A | N/A | N/A | N/A | N/A |
| Hakim, Mahmud and Morsin 2024 [54] | AutoTFT | 12 (ST1D) | GLC | N/A | N/A | 7.71 | N/A | 8.44 | N/A | N/A | N/A | N/A | N/A |
| Shuvo et al., 2024 [55] | Deep-LSTM | 6/6 (O18+O20) | GLC | 2 h | N/A | N/A | N/A | 31.96 | N/A | 47.67 | 58.08 | N/A | N/A |
| Wang et al., 2024 [56] | WaveNet | 12 (O18+O20) | GLC, insulin, PA, CHO | 2h | N/A | 13.70 | N/A | N/A | N/A | N/A | N/A | N/A | N/A |
| Yang et al., 2024 [57] | GRU + TCN | 12 (O18+O20) | GLC, insulin, CHO, PSY data | N/A | N/A | 16.90 | N/A | 28.88 | N/A | N/A | N/A | N/A | N/A |
| Zheng, Ji and Wu, 2024 [58] | BG-BERT | 12/54 (O18+O20/ DT) | GLC | 2h - 4h | N/A | 14.02/ 14.85 | N/A | 23.67/ 24.95 | N/A | N/A | N/A | N/A | N/A |
| Giancotti et al., 2024 [59] | GRU | 6/ 17 (O20/ PGC) | CGM, HR | 1 h/ 2 h | 12.39 / 5.57 | 20.48/ 9.30 | N/A | 34.16/ 16.15 | N/A | N/A | N/A | N/A | N/A |
| Nemat et al., 2024 [60] | SVR | 6/6 (O18/ O20) | CGM, insulin, CHO, activity | N/A | N/A | 19.59/ 19.77 | N/A | 31.92/ 32.95 | N/A | N/A | N/A | N/A | N/A |
| Shen and Kleinberg, 2025 [61] | Stacked LSTM | 126 (OAPSSK)/ 226 (RBG) | CGM | 2 h | N/A | 10.23/ 13.41 | N/A | 15.08/ 19.68 | N/A | N/A | N/A | N/A | N/A |
| Neumann et al., 2025 [62] | LSTM | 79 (T1DEXIN) | CGM, insulin, PA, HR, CHO | 2 h | N/A | 18.21 | N/A | N/A | N/A | N/A | N/A | N/A | N/A |
| Piao et al., 2025 [63] | GARNN | 12/12/12 (O18+ O20/ A/ ST1D) | CGM, PHY, PSY, insulin, CHO | 4 h | N/A | 18.97/ 19.97/ 13.62 | N/A | N/A | N/A | N/A | N/A | N/A | N/A |
| Xiong et al., 2025 [64] | TF+ BiLSTM | 13 (YT1D) | CGM, insulin, meal, personal features | 4 h | N/A | 10.16 | N/A | 10.64 | 13.34 | 13.97 | N/A | N/A | N/A |
| Mazgouti et. al, 2025 [65] | LSTM+ XGB | 12 (PMZ) | CGM | N/A 14 days | 7.97 | 9.63 | 10.72 | 10.93 | N/A | N/A | N/A | N/A | N/A |
| Kalita and Mirza, 2025 [66] | MLP+ MHAL | 6/6 (O18/ O20) | CGM, CHO, insulin, HR, activity | N/A | 5.78/ 6.55 | 16.57/ 18.27 | N/A | 29.25/ 29.73 | N/A | N/A | N/A | N/A | N/A |

*CCMBA, Convergent Cross Mapping-Based Approach. CHO, Carbohydrate. CNN, Convolutional Neural Networks. CRNN, Convolutional Recurrent Neural Networks. DT, Decision Tree. FFNN, Feed Forward Neural Networks. GLC, Glucose. GPCEM-PSO-BP, Glucose Prediction algorithm combined Correlation coefficient-based complete ensemble empirical mode decomposition with adaptive noise - Particle swarm optimization - and back propagation neural network. GRU, Gated Recurrent Unit. ISL, Input Sequence Length. LMU, Legendre Memory Unit. LR, Linear Regression. MARS, Multivariate adaptive Regressions Splines. MLP, Multilayer perceptron. N/A, Not Available. NNARX, nonlinear auto-regressive with exogenous input neural network. PA, Physical Activity. PBGTAM, prediction method of blood glucose based on temporal multi-head attention mechanism. PH, Prediction Horizon. RF, Random Forest. RL, Reinforcement Learning. SOM, Self-Organizing Map. SVR, Support Vector Regression. TCN, temporal convolutional networks. WOA, Whale Optimization Algorithm. XGB, eXtreme Gradient Boosting.*

*Table 2: Classification Models for Short- and Long-term PHs*

| Study | Metric | Model | Input Data | Input Features | ISL | Performance of PHs | | | | | | | |
|---|---|---|---|---|---|---|---|---|---|---|---|---|---|
| | | | | | | Short-term PH in min | | | | | Long-term PH in h | | |
| | | | | | | 15 | 30 | 45 | 60 – 75 | 120 | 3-4 | 6-9 | 24≤ |
| Gadaleta et al., 2019 [16] | REC, PRE, F1 (level 1/ level 2) | SVM | 89 (PC) | GLC | 25 min | N/A | 0.86, 0.36, 0.49/ 0.98, 0.32, 0.47 | N/A | N/A | N/A | N/A | N/A | N/A |
| Oviedo et al., 2019 [67] | SE, SP (level 1/ level 2) | SVC | 10 (PO) | GLC, insulin, CHO | 30 min - 2h | N/A | N/A | N/A | N/A | N/A | 0.71, 0.79/ 0.77, 0.81 | N/A | N/A |
| Vehí et al., 2019 [68] | SE, SP (level 1/ level 2) | SVM | 10 (PO) | GLC, insulin, CHO | 30 min - 4h | N/A | N/A | N/A | N/A | N/A | 0.69, 0.80/ 0.75, 0.81 | N/A | N/A |
| Vehí et al., 2019 [68] | SE, SP | MLP | 6 | GLC, insulin, PA | 6 h | N/A | N/A | N/A | N/A | N/A | N/A | 0.44, 0.86 | N/A |
| Bertachi et al., 2020 [69] | SE, SP, AC | SVM | 10 (PB) | GLC, PHY, sleep, insulin, CHO | 6 h | N/A | N/A | N/A | N/A | N/A | N/A | 0.78, 0.82, 0.81 | N/A |
| Dave et al., 2020 [70] | SE, SP | RF | 112 (PD) | GLC, insulin, CHO | 1 – 4h | 0.98, 0.98 | 0.97, 0.95 | 0.97, 0.95 | 0.96, 0.96 | N/A | N/A | N/A | N/A |
| Jensen et al., 2020 [71] | SE, SP | LDA | 463 (NN) | GLC, insulin, CHO, static data, BMI | 3 days | N/A | N/A | N/A | N/A | N/A | N/A | 0.75, 0.70 | N/A |
| Vu et al. 2020 [72] | AUC | RF | 9800 (PV) | GLC | 6h | N/A | N/A | N/A | N/A | N/A | 0.90 | 0.84 | N/A |
| Berikov et al., 2022 [73] | SE, SP, AUC | RF | 406 (RICEL) | GLC, static data, lab data | 1h | 94.5, 91.4, 0.97 | 90.4, 87.4, 0.94 | N/A | N/A | N/A | N/A | N/A | N/A |
| D'Antoni et al., 2022 [74] | REC, PRE, F1 | LSTM-ME-DT | 12 (O18 +O20) | GLC | 30 min | 0.81, 0.77, 0.79 | 0.55, 0.69, 0.61 | N/A | 0.29, 0.45, 0.35 | 0.26, 0.22, 0.22 | N/A | N/A | N/A |
| Fleischer, Hansen, and Cichosz, 2022 [75] | AUROC / PR-AUC/ SE | RUS-Boost | 225 (RBG) | GLC | 1 h | N/A | 0.99/ 0.77/ 0.90 | N/A | N/A | N/A | N/A | N/A | N/A |
| Parcerisas et al., 2022 [76] | SE, SP, F1 | SVM | 10 (PB) | GLC, insulin, CHO, PHY, sleep | 6 h | N/A | N/A | N/A | N/A | N/A | N/A | 0.74, 0.76, 0.76 | N/A |
| Shastri and Sandhya, 2022 [77] | AC | RF | 70 (UCI2005) | GLC, insulin, CHO | N/A | N/A | N/A | N/A | 0.82 | N/A | N/A | N/A | N/A |
| Shastri and Sandhya, 2022 [77] | SE, SP | ANN | 6 | GLC, insulin, CHO, PA | 6h | N/A | N/A | N/A | N/A | N/A | N/A | 0.85/ 0.81 | N/A |
| Alvarado et al., 2023 [78] | AC | CNN | 4 | GLC | 24h | N/A | N/A | N/A | N/A | N/A | N/A | N/A | 0.79 |
| Felizardo et al., 2023 [79] | SE, SP | RF + SkNN | 54 (UCI) | GLC, insulin, CHO, PA | 24h | N/A | N/A | N/A | N/A | N/A | N/A | N/A | 0.45, 0.89 |
| Kozinetz et al., 2023 [80] | F1 | MLP | 380 (RICEL) | GLC | 30 min | N/A | 0.80–0.86 | N/A | N/A | N/A | N/A | N/A | N/A |
| Mosquera-Lopez et al., 2023 [81] | AUROC | MERF | 50/20 (TP2/ T1Dex) | GLC, insulin, PA, static data | 24h | N/A | N/A | N/A | 0.83/ 0.86 | N/A | N/A | N/A | 0.66/ 0.68 |
| Piersanti et al., 2023 [7] | SE, SP, F1 | DT | 50 (DN05) | GLC, static data | 2 – 5 h | N/A | N/A | N/A | N/A | N/A | N/A | N/A | 0.87, 0.76, 0.87 |
| Giammarino et al., 2024 [82] | F1 | linear classifier | 226/ 355/ 120 (RBG/ JDRF/ TP1) | GLC | 1 week | N/A | N/A | N/A | N/A | N/A | N/A | N/A | 0.75/ 0.66/ 0.68 |
| Cichosz, Olesen, and Jensen, 2024 [83] | ROC-AUC | XGB | 187/ 223 (WISDM, CITY+ RBG)) | GLC, static data | 1-4 weeks | N/A | N/A | N/A | N/A | N/A | N/A | N/A | 0.89/ 0.89/ 0.84 |
| Leutheuser, 2024 [84] | F1, F2, AUC | RF | 13 (PM) | GLC, PHY | N/A | N/A | N/A | N/A | N/A | N/A | N/A | 60.9, 61.3, 75.2 | N/A |
| Leutheuser, 2024 [84] | F1, F2, AUC | LR | 13 (PM) | GLC, static data | 45 min | 43.3, 63.9, 90.7 | N/A | N/A | N/A | N/A | N/A | N/A | N/A |

*AC, Accuracy. ANN, Artificial Neural Networks. AUC, Area Under the Curve. AUROC, Area Under the Receiver Operating Characteristics Curve. BMI, Body Mass Index. CHO, Carbohydrate. CNN, Convolutional Neural Networks. DT, Decision Tree. F1, F1-Measure. F2, F2-Measure. GLC, Glucose. ISL, Input Sequence Length. LDA, Linear Discriminant Function. MERF, mixed-effects Random Forest. MLP, Multilayer perceptron. N/A, Not Available. PA, Physical Activity. PH, Prediction Horizon. PRE, Precision. REC, Recall. RF, Random Forest. RUSBoost, Random Under Sampling Boosting. SE, Sensitivity. SkNN, Subspace k-Nearest neighbor. SP, Specificity. SVC, Support Vector Classification. SVM, Support Vector Machines. XGBoost, eXtreme Gradient Boosting*

*1. PH of 15 min*

13 studies forecast 15 min, with most of them utilizing DL, EL, or hybrid models. ML models are only leveraged by Georga et al. [24] and Zhu et al. [39]. However, only Georga et al. achieved sufficient RMSE values of 5.12 mg/dL with an individual-based (IB) Support Vector Regression (SVR) model. Studies using DL models report better performance. RMSE values below 5 mg/dL are achieved by Annuzzi et al. with 4.41 mg/dL and 2.91 mg/dL [40], Alfian et al. with 2.82 mg/dL [28], Allam (2021) with 2.56 mg/dL (0.14 mmol/L) [30], and lastly, Nasir et al. with 2.23 mg/dL [44]. Comparing the methods, Alfian et al. train individualized Multilayer Perceptrons (MLPs) [28], Annuzzi et al. and Allam utilize Feed Forward Neural Networks (FFNN), whereas Allam further expands the model with a whale optimization algorithm (WOA) [30,40]. Nasir et al. build a two-layered sequence-to-sequence LSTM model [44]. As input data, all studies except Allam choose an ISL of 30 min from 12 children. Allam leverage the last 50 min from 9 subjects, but more information is not given. The second dataset of Annuzzi et al. is from 25 adults and includes nutritional features. The other studies use just CGM data, and also Alfian et al. include time-based features.

*2. PH of 30 min*

A PH of 30 min is reported by 36 studies. Mazgouti et al. [65], Giancotti et al. [59], Jaloli and Cescon [33], Katsarou et al. [34], Annuzzi et al. [40], Hakim, Mahmud and Morsin [54], Ben Ali et al. [26], and Nasir et al. [44] obtain a similar RMSE value less than 10 mg/dL with 9.63, 9.30, 9.28, 8.81, 8.30, 7.71, 7.45, and 7.38 mg/dL, respectively. Mazgouti et al. present a population-based (PB) hybrid model of XGBoost (XGB) and LSTM networks. They train the model on CGM values of 12 patients [65]. Ginacotti et al. utilize a GRU with a dual attention mechanism train on CGM and HR data of 6 patients of the OhioT1DM 2018 dataset and test on 17 private patients. While the model only achieves an RMSE score of 20.48 on the trained dataset, a score of 9.30 mg/dL is reported for the test dataset [59].
Jaloli and Cescon develop a hybrid CNN-LSTM network and train on 168 subjects using 60 min of past CGM, insulin, and carbohydrate (CHO) data as input. A further validation on 29 subjects indicates generalization with a variation of 0.53 mg/dL [33]. Katsarou et al. train personalized LSTM models and extract the glucose profile including daily glucose graphs, statistical, and demographic data from 29 patients. The RMSE value reported for a 15-min PH shows an increase of 2.4 mg/dL for 30 min [34]. Hakim et al. [54] apply auto-tuned temporal fusion transformers (AutoTFT) and Ben Ali et al. [26] report an FFNN trained with a Levenberg Marquardt algorithm and show a variation of 1.02 mg/dL between 15 and 30 min. Both train individual-based models from the CGM data of 12 subjects. Looking at the stability of previously reported studies, Anuzzi et al. have an increase of 4.14 and 5.13 mg/dL [40]. Nasir et al. show variations of just 0.15 mg/dL, but the robustness cannot be identified since more than 30 min are not forecasted [44].
Better results are obtained by Allam (2021) [30], Munoz Organero [29], Alfian et al. [28], and Georga et al. [24] with RMSE values between 6.49 [30] and 6.03 mg/dL [24]. While the performance is similar, different algorithms,

ISL, and datasets are used. Munoz Organero tests a virtually trained model on 9 real subjects. A mathematical model is combined with an LSTM-RNN network to simulate the individual metabolic process of glucose inputting data of CGM, insulin, and meal intake [29]. Again, Georga et al. present the only ML model with sufficient performance. A Support Vector Regression (SVR) is trained integrating expended energy during activity, and a mathematical insulin and meal processing model. 15 out of 25 subjects have data collected of CGM glucose, insulin, carbohydrate (CHO) and PA and achieve better results than the groups with less features. For 15 minutes, an RMSE value of 5.21 is reported [24]. Alfian et al. and Allam, reported previously, show a moderate increase of 3.49 [28] and 3.93 mg/dL (0.22 mmol/L) for 30 min, respectively. An ISL of 30 min is leveraged by Georga et al. and Alfian et al. [24,28], of 60 min by Allam [30], and of 9 hours by Munoz Organero [29].
Lastly, the least RMSE value of 1.425 mg/dL is described by Yang et al. (2023). They train an individual-based adaptive weighted deep ensemble learning (AWD-stacked) model on glucose data from 12 subjects. The base-estimators of the meta-learner consist of a 3-layered StackedLSTM, a VanillaLSTM, and a BiLSTM, and the combiner is a decision tree (DT). Finally, the meta-model consists of a linear regression (LR). The base estimators are adaptively weighted with a weighted similarity matrix [49].

*3. PH of 45 min*

Only 12 studies forecast 45 min. Annuzzi et al. [40], Allam (2021) [30], Allam (2024) [51], Alfian et al. [28], Mazgouti et al. [65], and Ben Ali et al. [26] report RMSE values less than 15 mg/dL with 13.72, 13.51, 13.42, 10.72, 10.65, and 8.13 mg/dL, respectively. They have variations of 5.42, 7.02 (0.39 mmol/L), -3.01 (-0.17 mmol/L), 3.34, 1.66, and 0.68 mg/dL, respectively, from 30 to 45 min. Consequently, Ben Ali et al., Mazgouti et al., and Alfian et al. present more consistent values, whereas Allam (2021) has the greatest variation. With a follow-up study, Allam investigates the robustness of a nonlinear autoregressive exogenous input neural network (NNARX) for longer short-term PHs. The model is a modified version of a prior best RNN architecture featured in [85]. It shows a lower STD for all PHs, and a better RMSE value for 45 and 60 min. However, the values for 15 and 30 min are significantly increased, indicating that different PH may require different approaches [51].
Again, the method of Yang et al. is superior, with RMSE values of 3.21 mg/dL, showing an increase of 1.79 mg/dL. Since 15 min are not forecasted, it cannot be assumed to be the best approach for forecasting 15 to 45 min of glucose level [49]. ML-based models report increased RMSE scored of more than 19 mg/dL.

*4. PH of 60-75 min*

Coming now to a PH of 60 min, 35 studies are presented, of which 3 utilize ML models but again, only Georga et al. report sufficient performance. In contrast, 7 studies using DL models reach an RMSE value less than 20 mg/dL. However, only Allam (2024) [51], Munoz Organero [29], Alfian et al. [28]  Xiong et al. [64], and Mazgouti et al. seem promising to detect adverse events on time with 15.33, 13.54, 11.35, 10.64, and 10.93 mg/dL, respectively [65]. Martinssons et al. present an RMSE value of 18.87 mg/dL, but still interesting findings are observed since explainability is investigated with an RNN model [27]. Results of previously reported approaches vary with increasing PHs. Xiong et al. combine Transformer and BiLSTM networks, which work in parallel, to capture long-term and short-term dependencies. They train on patient data, insulin and meal information, and CGM values as input. They also forecast 30 min of glucose values showing a very small increase of 0.48 mg/dL to 60 min [64]. Mazgouti et al. report the most stable results with an increase of only 0.21 mg/dL [65]. Annuzzi et al. still obtain good performance with an increase of 2.97 mg/dL, whereas Alfian et al. report an increase of 4.68 mg/dL from 45 min to 60 min. Munoz Organero shows an increase of 4.93 mg/dL from 30 to 60 min.
Ben Ali et al. [26], Hakim et al. [54], Georga et al. [24], Yang et al. [49], and Rajagopal and Thangarasu [45] report RMSE values less than 10 mg/dL with 9.03, 8.77, 7.14, 6.35, and 6.30 mg/dL, respectively. From 30 to 60 min, minimal variations are noticed in Hakim et al. (0.73 mg/dL) and Georga et al. (1.09 mg/dL). Ben Ali et al. and Yang et al. report variations of 0.90 and 3.13 mg/dL, respectively, from 45 to 60 min. Rajagopal and Thangarasu use features extracted with autoencoders as input data to the LSTM models. They train on 12 subjects, of whom CGM, insulin, and CHO intake were collected. Furthermore, 75 min are forecasted with an RMSE value of 8.0 mg/dL.
Finally, the best performance can be seen in Ramachandran et al. with an RMSE score of 4.12 mg/dL [46], and Bi and Kar [9] of 3.86 mg/dL. Ramachandran et al. develop an individual-based 2DCNN-LSTM network which is trained on 12 subjects. They leverage CGM data, bolus insulin doses, and CHO intake [46]. Bi and Kar, similar to Yang et al., build a stacked model called the Balanced multi-model scheme (BUMS). The base-estimators consist of LSTM, RF, and XGBoost models, of which the outputs are forwarded to a pretrained balancer based on LR. The balancer weights the predictions so that more certain models have more impact. As input, CGM data from 25 children are employed, whereas the balancer is trained with data from 5 children. In comparison to other approaches, not the exact PH is forecasted but a horizon range of 30-60 min. In general, glucose values are reported in mg/dL, but the metric of the RMSE value is not directly mentioned. Nevertheless, the presented graphs indicate that the results are measured in mg/dL. Conclusively, it is highlighted that multi-model systems use the advantage of each model and could mitigate limitations of single models [9].

*5. PH of 80-90 min*

8 studies predict glucose values 80- 90 min in advance, of which none use ML models. An RMSE value less than 25 mg/dL is reported by Allam (2024) with 23.45 mg/dL [51], Jaoli and Cescon with 18.40 mg/dL [33], and Xiong et al. with 13.34 mg/dL [64], revealing increases of 4.86, 6.94, and 2.7 mg/dL, respectively. Rajagopal and Thangarasu [45] obtain an RMSE value less of 9.40 mg/dL with an increase of 3.1 mg/dL. Finally, Cai et al. achieve a value of 6.46 mg/dL. They propose a correlation coefficient-based complete ensemble empirical mode decomposition with adaptive noise and back propagation neural network (GPCEMBP). CGM glucose is decomposed into multiple time-frequency components, the noise is filtered with correlation adaptive screening, and finally, the glucose value is reconstructed. The model is trained and validated on adult data and additionally validated on 3 children of another dataset [41].

*6. PH of 120 min*

9 studies are presented for a PH of 120 min. Xiong et al. report RMSE values of 13.97 mg/dL, increasing by 0.63 mg/dL compared to a 90 min PH. Thus, they obtain very stable results across prediction horizons with the same model and dataset [64]. Best performances are noticed in Georga et al. with 7.62 mg/dL using a history window of 90 min and Bi and Kar with 4.39 mg/dL. Additionally, both show less variation to shorter PHs and, after forecasting, identify hypoglycemic events. Georga et al. define hypoglycemia with a threshold of 60 mg/dL, and Bi and Kar with 70 mg/dL. An accuracy of 87%, 83%, and 85% for 30, 60, and 120 min, respectively, is illustrated by Georga et al. Additionally, Bi and Kar report that 26 out of 27 events are detected, but the PHs are not given. Other studies are clinically insufficient with RMSE values of at least 30 mg/dL.

*7. PH of more than 2 h*

5 studies forecast long-term PHs which are Tyler et al. [37] and Shuvo et al. [55] with 4h, Phadke and Nagaraj [36] with 3 h, Bi and Kar [9] with horizons from 2-4, 4-8, and exactly 8h, and Zafar et al. [50] with 48h.
Tyler et al. study exercise-induced hypoglycemia and predict 40 min and 4 h with an RMSE value of 19.7 mg/dL, increasing to 23.0 mg/dL. They train an individual-based Multivariate adaptive Regressions Splines (MARS) model on CGM data, data of PA, and historical exercise information from 20 subjects. After forecasting, the risk for hypoglycemia is predicted with a sensitivity of 73%, a specificity of 95%, and an accuracy of 88% for a 40-minute PH. For a 4-hour PH, the sensitivity decreases to 56%, whereas the specificity and accuracy remain similar. The population-based model for 40 min achieves similar results. Contrariwise, for 4 h, a better sensitivity by 23% is obtained, whereas other metrics are decreased [37]. Shuvo et al. forecast 1, 2, and 4 h with a deep LSTM. They employ multi-task learning (MTL) and leverage the last 2 hours of CGM data from 12 subjects. Reported RMSE values are 31.96, 47.67, and 58,08 mg/dL, respectively, which seems insufficient for clinical use cases. The Clarke error grid shows that the predictions are mainly within clinically acceptable regions but not in the safe zone [55]. Studies applying only ML models, do not focus on PHs longer than 4 hours.
Phadke and Nagaraj train an ANN utilizing data from 10 subjects and focus on PHs of 30 min, 1, 2, 3, and 24 h. The corresponding mean absolute percentage errors (MAPE) are 0.037, 0.069, 0.149, 0.215, and 0.134, respectively. The input features include CGM data, insulin, CHO intake, PA, physiological (PHY), and psychological (PSY) signals. They are the only researchers predicting up to 24 h before, leveraging a multivariate dataset with PHY and psychological parameters. The 24 h horizon is split into seven time-flag buckets, which are later realigned. A comparison with other models shows that ANN performs better for long-term predictions [36].
Bi and Kar forecast up to 8 h with an RMSE value of 5.08 mg/dL. In total, they predict from 30-60 min, 1-2 h, 2-4 h, 4-8 h, and exactly 8 h with RMSE values of 3.86, 4.39, 4.99, 5.10, and 5.08 mg/dL, respectively. It is reported that the ensemble-meta-learner is better in all PHs than using single algorithms. However, for exactly 8 h, RF has decreased RMSE values of 4.81 mg/dL. Thus, it is again highlighted that different PHs may need different methods. If the reported values are in mg/dL, the model shows good stability across short-term and long-term PHs with less variation between [9].
Finally, Zafar et al. predict a 48-h glucose profile with an RMSE value of 3.7 mg/dL. They train an individual-based LSTM model on 19 subjects, inputting the glucose profile of 60 days. Glucose profiles are extracted from CGM data, descriptive statistics, and demographics [50]. They report even better performance than Bi and Kar and use a single classifier. The main identified difference is the utilized dataset, containing 96-1688 days per subject. This study duration exceeds the average length of datasets used in regression methods.

*8. Impact of Dataset, Features, and Settings*

Approaches cannot be compared directly, since different datasets and input features are utilized. Hence, the best-performing model could behave worse in unseen data. Details of the datasets can be seen in section III C. For regression, usually the OhioT1DM dataset is chosen, yielding similar average results. Only Yang et al. report outstanding performance. 6 out of 44 studies train on children's data, and the most popular is the DN07 dataset, which typically has better results than the average. Marx et al. collect their data from children spending time at a supervised sports camp. They obtain the least significant performance, which could be due to the irregular data [43]. Consequently, the impact of data quality and superior approaches is recognized. Fig. 4 illustrates that 70% of utilized databases contain only up to 20 subjects, and 16% contain 20-60 subjects. Larger datasets of more than 70 participants are only assessed by 7 studies (14%), representing a better population, but average performance.

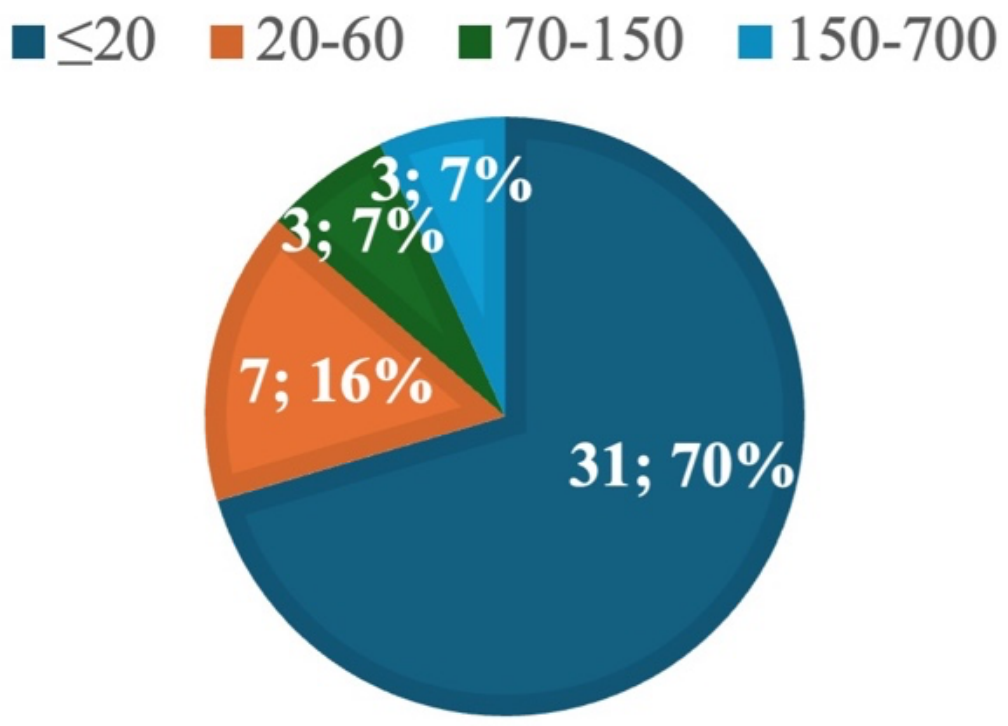

*Figure 4: Distribution of Utilized Dataset Sizes by Regression Models. The chart shows the proportion of leveraged dataset sizes. Each slice corresponds to the number and percentage of studies covering the proposed dataset sizes. For studies using multiple datasets, the number of subjects is summed up.*

In general, individual-based models perform better and are advantageous if more data from one subject is available [50]. When Annuzzi et al. test on data of children, individual-based models have decreased values but with more variation than the population-based approaches [40]. In contrast, Jaloli and Cescon present significantly increased RMSE values and STD for individual-based models [33]. Usually, from the reported results, better performance was obtained with population-based models. However, it cannot be directly inferred if individual or population-based methods lead to better performance since the results depend on the validation scheme, data volume, data quality, and heterogeneity. Furthermore, testing different training methods, Neuman et al. present that population-based models, which are later fine-tuned on data of a specific patient for personalization, perform best for most of the participants [62].

In total, 8 studies utilize multiple datasets for validation, exploring transfer learning, proving stability, or showing that individual models are better [33,39,41,53,58]. Ghimire et al. leverage 4 different datasets and aim to find a model performing best across multiple datasets with different age groups. The OhioT1DM and the RT datasets present more variation between subjects. However, the models perform best on the OhioT1DM dataset [53]. Giancotti et al. compare two datasets with different glucose sampling rates and report that the dataset with a frequency of 15 min led to significantly better results, which may also be impacted by other factors [59]. Also, Cai et al. test adult and children's data on the same model, but the number of participants is not representative [41]. Lastly, Piao et al. present performance differences between the OhioT1DM, ARISES, and ShanghaiT1D datasets, while the ShanghaiT1D dataset performs the best [63].

Table 1 shows that 25 studies use only glucose, and 28 studies experiment with multivariate data. The inclusion of insulin and carbohydrate features is reported to improve performance [24,38,46,48] and is covered by 48% of studies. Additionally, finger-prick glucose values [39,41,47,58], sleep information [36,58], and wristband data such as inter-beat-interval (IBI) or heart rate variability (HRV) are suggested [39]. Giancotti et al. report that HR data improves performance [59]. Also, PA should increase performance and stabilize the model, particularly for hypoglycemia detection [24,25]. Including static features would be impactful for short-term predictions [28,43], and glycemic variability for a PH of 120 min [32]. The combination of physiological and psychological features like stress data would also result in improvement [36]. Finally, Annuzzi et al. reveal that proteins and fibers are effective features for 30 min, but not contributing to 15 min. All tested nutritional factors would improve the performance after 30 min if not used altogether. They suggest choosing individual-based nutritional features [40]. Additionally, Tyler et al. suggest that long-term PHs benefit from personal data [37]. Conclusively, only CGM data may be appropriate for shorter short-term PHs, and multivariate models may improve the performance for longer PHs [24,36].

As additional model architectures, Nemat et al. report that including causality as input measured with a convergent cross mapping (CCM) is beneficial. They show that the causation strength of activity is higher than that of carbohydrate, which has a stronger causation than bolus insulin [86]. Yang et al. (2024) highlight that building individual models on clusters of age and sex is effective [57]. Domanski et al. join LSTM and deep reinforcement learning (RL) [42]. Zhu et al. propose an attention-based bidirectional GRU model defined as a Fast adaptive and confident neural network (FCNN) and assert that model-agnostic meta-learning (MAML) improves performance with limited training data [38,39]. Wang et al. report a WaveNet model with three domain specific constraint mechanisms to provide counterfactual explanations. Generating counterfactual samples for hypoglycemia patients would be more challenging [87]. Zheng, Ji and Wu train a BG-BERT model with a Synthetic Minority Over-sampling Technique (SMOTE) data augmentation and a shrinkage loss function. Altogether, the architecture would compensate for the imbalanced data regarding hypoglycemia. The predictor consists of LSTM, CNN, and MLP models [58]. As already identified, ensemble learning could lead to better outcomes. Floris and Vasata propose Legendre Memory Units (LMU), applying two compartmental insulin and meal models to convert discrete data into continuous values [52]. Ghimire et al. testing the same model with various datasets, report that LSTM and self-attention networks (SAN) generate the best results with better generalization capability on unseen datasets [53]. Piao et al. propose a multivariate hybrid model of Graph Attention Networks (GAT) and GRU layers as a Graph Attentive Neural Network (GARNN). The GAT layers are used to model correlations across features and extract context from important connections, whereas RNN layers learn temporal features [63]. Kalita et al. report that Multi-Head Attention Layers (MHAL) increase performance [66].

Moreover, data normalization [26,42,53], smoothing, and cleaning are suggested, for instance with a Kalman filter [61], a Gaussian function [47], or a Savitzky-Golay technique [28,40]. Yang et al. (2023) report that data smoothing is advantageous for the PH of 30 min. They build a temporal multi-head attention mechanism that selectively keeps only relevant information of global and local features [48]. Yang et al. use Kalman filtering for addressing errors in CGM and double exponential smoothing. [49], MTL is suggested for generalization and performance improvement in small datasets, especially when using DL [32,47]. As already seen, Marx et al. and Bi and Kar demonstrate that different PHs benefit from different models [9,43]. Thus, the RNN model is better for 60 and 120 min, whereas the LSTM model is better for a 30 min PH [43].

Lastly, coming to the ISL, most studies consider 30 min or 2-4 h for short-term prediction horizons. Some adjust the ISL to the selected PH from 30 to 90 min, while Jaoli and Cescon use an ISL 3 times the length of the PH [24]. Shuvo et al. and Ghimire et al. use the past 2 h for multiple PHs up to 4 h [53,55]. Single studies choose a longer ISL of 9 h or 24 h [24,34]. Yang et al. (2023) investigate the effects of different history sequences and divide them into 30, 45, 60, and 90 min. Then, the average value is used as the final result [49]. Finally, Ben Ali et al. decide on individual ISL varying from 90 to 135 min per subject for a PH of 15 min. It is not known if the same ISL is utilized for all PHs [26].

## B. Summary of Classification Models

The featured studies for hypoglycemia classification are presented in Table 2. The distribution of the chosen PHs, shown in Fig. 2, reveals that 7 out of 20 studies select a PH of 6-9 h, thus classifying nocturnal hypoglycemia. 6 studies classify 30 min before, and 6 studies select a PH of at least 24 hours. In general, long- and short-term PHs are covered with the same percentage, whereas in total, 12 studies report long-term, and 7 studies short-term PHs. From Table 2, it can be observed that all studies but four, which classify long-term PHs, consider only one PH. Conversely, when focusing on short-term classification, the same model and data are used to classify across multiple PHs. The least covered PHs are 45 and 120 min. Looking at Fig. 5, 13 studies (62%) utilize conventional ML models, 4 studies (19%) use NN or DL, and the last 4 studies (19%) apply EL and hybrid models. Thus, conventional ML methods appear to either perform better or be more convenient with the selected data cohorts.

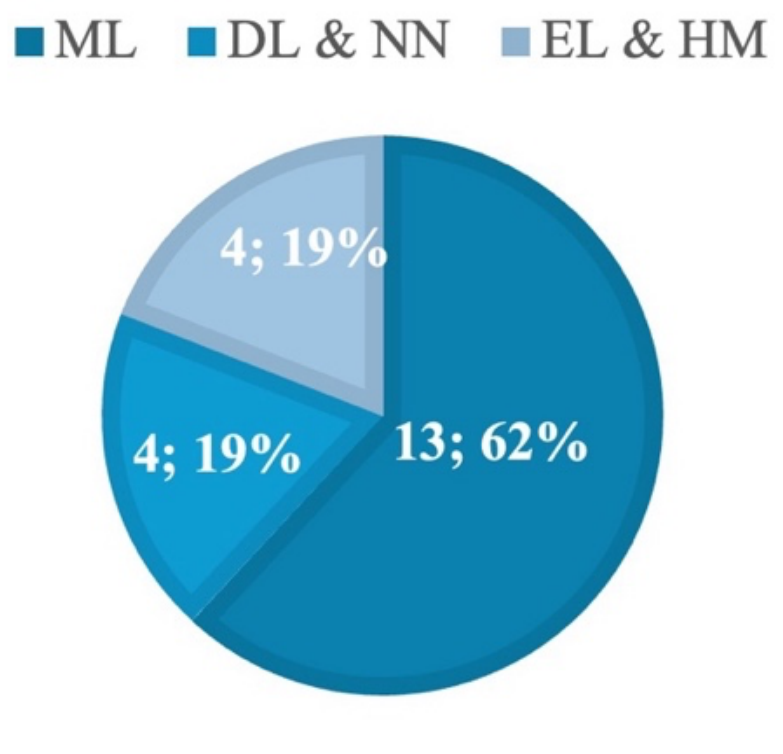

*Figure 5: Distribution of DL, EL, and ML Among Classification Studies. The chart presents the proportion of different machine learning architectures. Each slice corresponds to the number and percentage of studies covering the model architecture. DL, Deep Learning. EL, Ensemble Learning. ML, Machine Learning. NN, Neural Networks.*

In the following, the studies classifying hypoglycemia are compared. The studies are mainly differentiated in terms of continuous classification of a PH up to 2 h, in terms of postprandial-, nocturnal-, and exercise-induced hypoglycemia, and, finally, in terms of the risk assessment at least 24 h before. Then, the impact of the leveraged datasets, features, and model architectures is examined.

*1. PH of 15-120 min*

30% of studies classify up to 120 min prior to a hypoglycemic event. These can be divided into studies considering nocturnal glucose ranges, severity classification, and systems including hyperglycemia and hypoglycemia.

Fleischer, Hansen, and Cichosz classify 40 min before the onset, and present a model called RUSBoost (random undersampling) which should perform well on imbalanced data since it combines data sampling and boosting [88]. CGM data of 225 subjects is utilized with an ISL of 1 h. The hypoglycemic event is labeled with glucose values less than 70 mg/dL for at least 15 min. Sample- and event-based classification is explored, but the report of different metrics disables direct comparison. For sample-based classification, an area under the receiver operating characteristics curve (ROC-AUC) and precision-recall (PR)-AUC of 0.99 and 0.77 are reported, respectively. Event-based classification obtains a sensitivity of 90% and a false positive rate of 38% [75].

Berikov et al. assess nocturnal hypoglycemia with PHs of 15- and 30-min applying random forest (RF). The input features consist of features extracted from CGM glucose, laboratory data collected in hospitals, and static data such as demographics. A sensitivity, a specificity, and an AUC of 94.5%, 91.4%, and 0.97 for 15 min and 90.4%, 87.4%, and 0.94 for 30 min are obtained, respectively [73]. Leutheuser et al. test the same model with a different dataset and achieve worse results, whereas Logistic Regression (LR) is superior with a sensitivity of 90.4%, a specificity of 90.4%, and an AUC score of 90.7% [84]. Similarly, Kozinetz et al. focus on nocturnal hypoglycemia from 0:00 to 6:00 am. They differentiate between hypoglycemic, hyperglycemic, and euglycemic glucose ranges. Data of CGM glucose, insulin, and demographics from 380 subjects are leveraged. A population-based multi-layer perceptron (MLP) provides a precision, recall, and F1-measure of 87%, 86%, and 86%, respectively, for identifying hypoglycemia 30 min in advance with an ISL of 30 min [80]. Both studies use the same dataset, while Berikov et al. include more subjects and features and obtain increased performance. Moreover, Fleischer et al. achieve the same sensitivity but do not report specificity. Likewise, Dave et al. classify night-time hypoglycemia between 11:00 pm and 6:00 am up to 1 h before. However, as a difference, a RF model is leveraged. The utilized dataset contains 112 subjects, but only 19 have data of CGM, insulin, and carbohydrate collected, yielding better results. Input features mainly range from 1 to 4 h. A sensitivity of 98.35%, 94.04%, 96.92%, 91.21% and a specificity of 97.75%, 95.23%, 94.89%, and 95.73% for PHs of 0-15, 15-30, 30-45, and 45-60 min are presented, respectively. They develop a binary classification as to why the PHs are considered separately by the same model [70].

Shastri and Sandhya use a RF model as well but detect the risk for severe hypoglycemia in 1-h intervals. They report a maximum accuracy of 82%, and select the features of CGM glucose, basal insulin, irregular diet, and snack times per day for 70 subjects [77].

Gadaleta et al. classify hypoglycemia (level 1) and severe hypoglycemia (level 2) 30 min before the onset with a SVM trained on the prior 25 minutes of 89 subjects' CGM data. Reported recall, precision, and F1-measure performance are 86%, 36%, and 49% for level 1, and 98%, 32%, and 47% for level 2 hypoglycemia, respectively. The model considers hyperglycemia and hypoglycemia, which is underrepresented in comparison, as to why the precision might be insufficient [16]. This finding is also confirmed by D'Antoni et al. They built a population-based layered meta model consisting of three binary classification models trained to classify euglycemia, hyperglycemia, or hypoglycemia. Hypoglycemia is defined as a threshold of less than equal to 70 mg/dL, with the condition that the event should be at least a 10-minute in duration or consist of 3 datapoints. The model is trained sample-based but evaluated event-based, in which each event is at least 10 min long, whereas euglycemia is represented sample-based. The base estimators consist of LSTM models, and the meta-learner is based on decision trees with classification and regression trees (CART). For PHs of 15, 30, 60, and 120 min, a recall of 81.0%, 54.8%, 29.1%, and 26.3%, a precision of 76.6%, 69.1%, 44.7%, and 21.9%, and an F1-measure of 78.8%, 61.1%, 44.7%, 35.2%, and 21.7% is achieved, respectively. Meta-learning increased the F1-measure and could predict hypoglycemia on average with a time gain of 22.8 min. The study uses the data of 12 subjects and an ISL of 30 min. The performance decreases with increasing PHs, and only the classification of 15 min before shows good performance [74].

*2. PH of 4 h*

Oviedo et al. [67] and Vehí et al. [68] classify between level 1 (below 70 mg/dL) and level 2 (below 54 mg/dL) postprandial hypoglycemia with a PH of 4 h. Both use the same dataset containing 10 subjects.

Oviedo et al. classify hypoglycemia 4 h after the meal intake or bolus insulin dose, if no other meal was consumed. They utilize individual-based support vector classification (SVC) and develop two approaches for each severity level. A median sensitivity and specificity of 71% and 79% for level 1 and 77% and 81% for level 2 hypoglycemia are achieved, respectively. Postprandial hypoglycemia was labeled with 8 different schemes and tested for each subject, producing varying results. As input data, CGM values from 30 min to 1 h before the meal, 2 h of prior insulin doses, and time-domain features were chosen [67]. Likewise, Vehí et al. train a SVM. They utilize 30 min to 1 h of CGM data before the meal, 1 h of prior bolus, and 2 to 4 h of prior basal insulin. A population-based model is presented with a sensitivity and specificity of 69% and 80% for level 1 and 75% and 81% for level 2 hypoglycemia, respectively [68]. Oviedo et al. show a slightly better performance with a different set of features and individualized models.

*3. PH of 6-9 h*

Turning now to nocturnal hypoglycemia, which is covered by 22% of the studies, a PH and ISL of 6 h is used by most of the studies, while Leutheuser et al. [84] classify 9 h before.
Bertachi et al. train an individual-based SVM leveraging data of CGM glucose, insulin, carbohydrate, and activity from 10 subjects. Furthermore, physiological models for insulin, carbohydrate and PA are utilized. A sensitivity of 78.75%, specificity of 82.15%, and accuracy of 80.77% are presented [69]. Similarly, Parcerisas et al. apply the same model, dataset, and features. They compare different machine learning methods, including LSTM, while SVM performs best. A sensitivity and specificity of F1-measure of 74% and 77% are achieved, respectively, for a population-based model. The median specificity for individualized models decreases to 68% while the sensitivity remains the same [76].
Vehí et al. utilize an MLP and data from 6 subjects. From the given information, it is asserted to be the OhioT1DM 2018 dataset. Data of CGM glucose, insulin, carbohydrate, activity, and sleep information are inputted, in addition to physiological models for insulin and PA. Individual features are extracted for each subject. Reported population outcomes are 80.1%, 44.0%, and 85.9% for accuracy, sensitivity, and specificity, respectively. The imbalance of the dataset is compensated for with an adaptive synthetic sampling [68]. Shastri and Shandhya copy Vehí et al. and use ANN with most probably the same dataset and features, thus the performance is very similar [77]. Parcerisas et al. cannot outperform Bertachi et al., Vehí et al. have a better specificity, but the sensitivity is significantly decreased, as to why SVM may be better than simple ANN, NLP, and LSTM models for the long-term classification of hypoglycemia. Since most of the utilized datasets are of small size, neural networks may be unable to learn significant patterns, potentially resulting in an increased number of missed cases or overfitting.
Jensen et al. classify the risk of severe hypoglycemia during the night with a linear discriminant function (LDA) for 463 subjects. A night is defined from 00:01 to 05:59 am and is only considered if at least one meal and one bolus dose were given the evening before. The prior 3 h of glucose, and 6 h of bolus insulin and carbohydrate intake are leveraged as input in addition to demographic data. Furthermore, the glycemic variability percentage is computed. A ROC-AUC of 0.79 is obtained with a sensitivity of 75% and a specificity of 70% [71].
Vu et al. use the largest dataset with 9800 subjects and train a RF. One night is clustered into a full night, ranging from 0-6 am, an early night, from 0-3 am, and a late night from 3-6 am. Obtained values are AUC with 0.84, 0.90, and 0.75, respectively. It is reported that late-night classification had a significantly increased false negative rate. Data of CGM glucose and demographics are used from which statistical, count-based, and temporal features are extracted [72]. Vu et al. obtain a better performance than Jensen et al. and represent the most subjects, thus again models such as SVM and RF seem superior. Both train population-based models. Nevertheless, more metrics, in addition to ROC-AUC, are not presented, which disables a thorough comparison.
Finally, Leutheuser et al. also classify nocturnal hypoglycemia with a RF model but consider a PH of 9 h and use a dataset with children. The night is defined between 10:00 pm to 7:00 am. Using CGM glucose and physiological data yields the best performance of 61.3%, 60.9%, and 75.2%. Nevertheless, the performance is worse than that of the other studies, which could be impacted by the used dataset [84].
In summary, larger datasets produced worse results but represent a better population. Data collected for children is only utilized by Leutheuser et al., which could be more irregular compared to the data for adults.

*4. PH of 24 h*

Another field of interest is the risk assessment of hypoglycemia within the next 24 h, which was the focus of four studies. Piersanti et al. use population-based decision trees (DTs) to estimate the long-term prediction risk of exercise-induced hypoglycemia. They leverage CGM data from 50 children. The event is defined if within the next 24 h after exercising hypoglycemia with a threshold of 60 mg/dL is experienced. Obtained results are 85.5% (AUC), 87.2% (classification accuracy), 86.9% (precision), 87.2% (sensitivity), 76.1% (specificity), and 86.9% (F1-measure) [7]. Likewise, Mosquera Lopez et al. investigate the risk of exercise-induced hypoglycemia. They train on 50 and validate on an additional 20 subjects. The input data consists of CGM glucose, insulin, and activity, while the validation data consists of CGM glucose and activity. They built a mixed-effects RF (MERF). Hypoglycemia classification 1 h after the activity yields AUROC values of 0.83 and 0.86 for the first and second datasets, respectively. The performance decreases to 0.66 and 0.68, respectively, for classifying hypoglycemia 24 h after activity. The 24-h horizon is split into 1-h segments. Hence, the risk is analyzed hourly [81].
Alvarado et al. use only CGM data from 4 patients, not implying generalization. They utilize a transformer function to generate 866 images of the time series sequence and then apply a CNN model. The average classification accuracy of the validation data is 80%, while an accuracy of 78.78% is achieved for the test data. An accuracy of 88% is reported to be the best performance, while an accuracy of 73% is the worst [78].
Finally, Felizardo et al. classify 3 classes being no risk, risk for hypoglycemia defined as less than 75 mg/dL, and a hypoglycemic event defined as less than 70 mg/dL, with a consensus decision. Different combinations of classifiers are tested, and an ensemble model of RF and Subspace k-Nearest neighbor (SkNN) is superior. In all combinations, the specificity and false alarm rate improve compared to single classifiers, but no combination could obtain the best metrics for all subjects. They utilize a dataset of 54 patients and leverage information of CGM glucose, insulin, meal intake, exercise, time, and contextual data such as meal or activity behavior. For the reported results, only the better outcomes for 23 subjects are considered. The obtained average performances are 75.3% (accuracy), 45.4% (sensitivity), 89.4% (specificity), and 13.5% (false alarm rate). Overall, more than 70% of events are predicted for 53% of all participants [79].
Comparing all approaches, Piersanti et al. achieve the best performance, while Felizardo et al. have a better specificity but significantly decreased sensitivity. Furthermore, high variations between subjects are reported. Therefore, the approach does not apply to each individual [7,79]. All studies besides Alvarado et al. train on at least 50 subjects. It can be seen that most approaches use the last 24 h of data as input [78,79,81], while Piersanti et al. only report an ISL of 2 up to 5 h, indicating that longer ISL could be less representative [7]. In general, 70-80% of hypoglycemic events can be predicted 24 h in advance.

*5. PH of more than 24 h*

Lastly, Giammarino et al. assess the risk of a severe hypoglycemic event one week in advance, with a PH ranging from 1 to 7 days. They utilize three different datasets with 226, 355, and 120 subjects, respectively. The population model is trained separately on each dataset and tested on the other two datasets. Giammarino et al. apply a linear classifier, which is L1 and L2 regularized. An ISL of 1 week of CGM glucose measurements is chosen, which was transformed using the minimally random convolutional kernel transform (MiniRocket), since the same length of input data was not available for each person. The achieved test F1-measures for datasets 1, 2, and 3 are 75%, 66%, and 68%, respectively. The internal ROC-AUC scores are 0.77, 0.74, and 0.76, respectively. Therefore, the model performs better on the internal dataset. The average external ROC-AUC of all datasets is 0.74. The authors state that the prediction of 0-2 days in advance is most accurate, with 70–80% of all events being recognized. However, classifying multiple days in advance appears challenging [82].
Cichoz et al. predict a one-week-ahead hypoglycemia risk for elderly patients. They train an XGBoost with features extracted from CGM data and the baseline patient profile. Reported ROC-AUC and PR-AUC scores are 0.89 and 0.72, respectively. Moreover, validating on two external datasets, generalization is induced with ROC-AUC scores of 0.84 and 0.89, and PR-AUC scores of 0.73 and 0.68, respectively, for each dataset [83].

*6. Impact of Dataset, Features, and Settings*

Similar to regression models, a direct comparison is disabled since various variables impact the performance. The distribution of dataset sizes and the number of studies can be seen in Fig. 6. 43% of studies utilize small datasets with at most 20 participants, while datasets with 70-150 and 150-500 subjects account for 19% each. Collecting own data in small cohorts is likely more feasible than doing so for large databases. Table 2 presents that 29% of the studies leverage just CGM data, 21% incorporate CGM, insulin, and meal intake, and 25% of studies include physiological data such as heart rate (HR) or PA. Another 25% include static data like demographics. Compared to regression models, datasets utilized generally contain more subjects. Likewise, it is reported that individual-based models performed worse or not significantly better with small datasets [69,76]. Larger datasets can reduce the risk of overestimation, represent a broader population, but can contain more intra-subject variation. Hence, more often decreased performance is obtained. For instance, Dave et al. achieve the best performance with a smaller dataset, whereas Fleischer et al. and Berikov et al. report a sufficient but decreased sensitivity for a PH of 30 min with larger datasets. Since Berikov et al. and Dave et al. both apply RF and multivariate datasets, the features of the latter could have more impact, or larger datasets could lead to more variation. Berikov et al. leveraged a clinical dataset and report that demographic data, laboratory data, diabetes duration, and comorbidities are relevant features. Moreover, the risk of hypoglycemia would be positively related to the insulin dose, diabetes duration, and proteinuria [73]. According to Dave et al., the effects of insulin and carbohydrate data would be mainly effective for 30 and 60 minutes. They further report that using interval PHs improves sensitivity, whereas exact PHs enhance specificity when the PH is longer than 30 min [70]. In addition, Leutheuser et al. illustrate that specificity increases if only features extracted from glucose are leveraged. Best sensitivity is achieved when including static features, and best precision when including physiological features. Consequently, depending on which metric is more important, other feature combinations are more appropriate, or ensemble models could be utilized. In general, physiological data improved the F2-score for the ML models, and static data contributed to long-term PHs [84]. Similarly, other studies highlight the importance of static data and personalization [72,81]. Individual-based models could yield better outcomes in postprandial hypoglycemia classification [67], and Felizardo et al. further show that the model performance depends on the individual data [79]. Nevertheless, in more cases, individual models do not show a significant impact compared with population-based models [69,74,76]. 5 out of 20 studies use the output of mathematical insulin, carbohydrate, or activity models as input for ML, but their relevance is not examined [68–70,76,77]. Typically, multivariate models enhance performance, only D'Antoni et al. state that additional information slightly decreases it. Thus, depending on the complexity of the data and the use case, different settings should be explored. Leutheuser et al. also show that, depending on the dataset and PH, a different model is superior while all of their models were limited by dataset size [84].

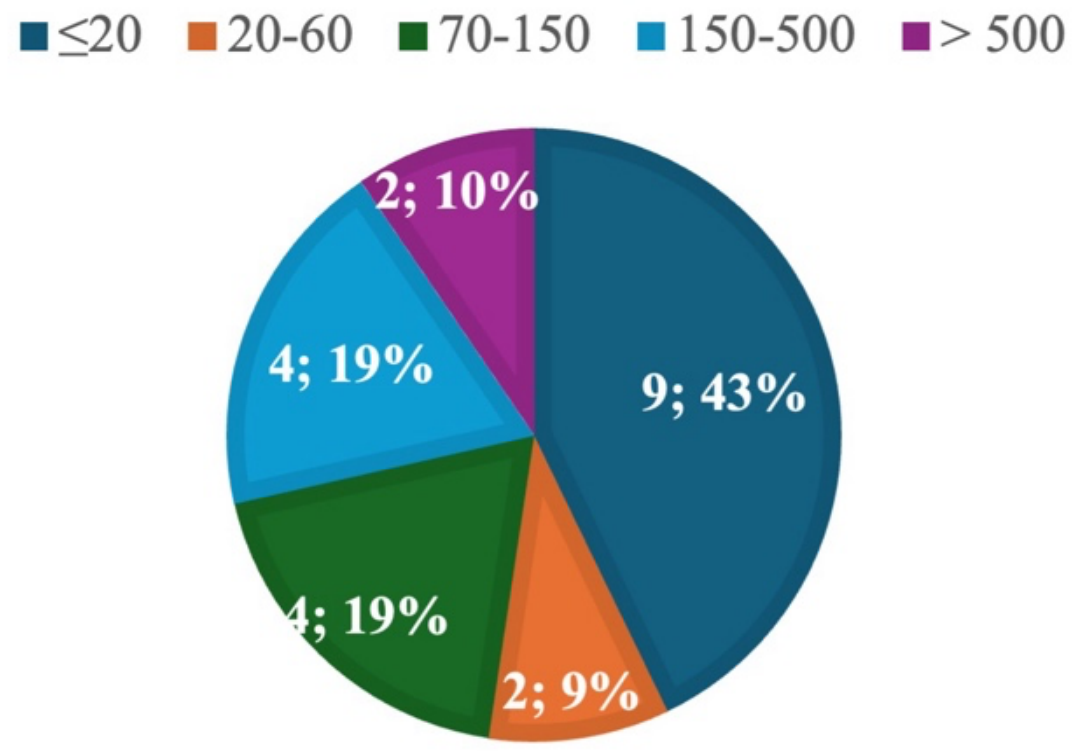

*Figure 6: Distribution of Utilized Dataset Sizes by Classification Models. The chart shows the proportion of leveraged dataset sizes. Each slice corresponds to the number and percentage of studies covering the proposed dataset sizes. For studies using multiple datasets, the number of subjects is summed up.*

Regarding nocturnal hypoglycemia, nighttime models yield better performance by 5% compared to datetime models [43,70]. Additionally, a difference between prediction performance in early night and late-night periods is reported by Vu et al. It is suggested to cluster subjects because subjects experiencing late-night hypoglycemia would have lower glucose values at midnight, an average downward glucose trend before midnight, a higher recent hypoglycemic event count, and no late-night hypoglycemia counts over the previous months [72]. Forward selection also shows that glucose data from the evening before nocturnal hypoglycemia and at bedtime, as well as the minimum glucose obtained from the previous night, are most predictive, including the BMI. In contrast, bolus insulin and meal intake are not relevant [71]. It is stated that basal insulin might be effective, which could not be investigated [71].

For exercise-induced hypoglycemia, risk factors such as glucose and body exposure to insulin at the start of the activity, low blood glucose index a day before the activity, and intensity and timing of the activity are identified. The greatest risk is identified as 1 h and 5–10 h after the activity [81]. The level of glucose deviation from the reference basal value, and a lower deviation level indicating better glycemic control, could be associated with hypoglycemia as well [7].

Coming to the impact of the model architectures, it can be observed that data preprocessing and model expansion are significant. Notably, when identifying underrepresented classes, balancing and sampling methods could be required, such as the RUSBoost algorithm [75]. However, these need to be validated on other real datasets to ensure that performance is not overestimated [75]. Conversely, Berikov et al. experiment with sampling methods, but report that neither undersampling nor oversampling resulted in great improvement [73]. D'Antoni et al. assert that sample-based classification can overestimate performance, which increases recall. However, sample- and event-based prediction are not directly compared. Another problem is that hypoglycemic datapoints are already underrepresented, and event-based classification would decrease the number of classified samples [74,75]. Felizardo et al. report that a consensus decision-based classifier using aggregated decisions by the majority vote yields better improvements than a single model. In particular, fewer false alarms are presented. Similar conclusions are noticed in D'Antoni et al., who use a layered meta-learning approach [74,79]. In addition, Mosquera Lopez et al. present that random effects impact accuracy for classifying hypoglycemia [81].

Looking at the ISL, different history sequences ranging from 25 min to 4 h are chosen for short-term classifications up to 120 min. Kozinetz et al. classified from 15 to 75 min before low glucose values and tested history sequences of 15 to 75 min, but report that the classification performance was not significantly affected. Thus, the optimal sequence length was decided to be 30 min. In general, a history sequence of up to 4 hours is used for postprandial hypoglycemia classification, while a 6-hour window is selected for nocturnal hypoglycemia classification.

Finally, a 24-hour window is commonly applied for a 24-hour risk classification. Subsequently, the ISL is usually as long as the PH for long-term PHs. Piersanti et al. present that shorter ISL leads to better results with a RF and decide on 2-5 h to classify the risk before 24 h.

## C. Datasets Utilized for Type 1 Diabetes

This section presents the datasets utilized in the reviewed studies. The datasets are grouped into small (Table 3), medium (Table 4), and large volumes (Table 5). Within each category, datasets are ordered by ascending number of subjects.

For small volume datasets comprising 6 to 20 subjects, it is observed that an increased number of features, especially wearable sensor data, are collected. However, the data collection duration is shorter compared to the medium-sized and large volume datasets. Additionally, variations in CGM measurement frequency are evident across studies, with some not reporting the sampling frequency. While the majority of small datasets are private, 7 out of 17 are available for free download or upon request. Moreover, only two studies collect data from children or teenagers. Data is usually measured under free-living conditions, without being required to follow a controlled diet or exercise regimen.

*Table 3: Small Volume Datasets Used by Presented Studies for Type 1 Diabetes*

| CT Number | Dataset | PN/ Age (Mean ± STD) | Study Duration | Description | Availability | CGM Frequency | Used by |
|---|---|---|---|---|---|---|---|
| N/A | OhioT1D 2018 (O18) [89] | 6/ 40-60 | 8 weeks | CGM, BGL from finger sticks, insulin dose, CHO intake, exercise, sleep and work times, reported stress and illness, hypoglycemic events, HR, step count, GSR, skin temperature, air temperature every 5 min | Free on request [90] | 5 min | [68,74] |
| N/A | OhioT1D 2020 (O20) [89] | 6/ 20-80 | 8 weeks | CGM, BGL from finger sticks, insulin dose, CHO intake, exercise, sleep and work times, reported stress and illness, hypoglycemic events, GSR, skin temperature, air temperature, MACC every 1 min | Free on request [90] | 5 min | [74] |
| N/A | Private by Cai et al. (PC) [41] | 7/ 24-55 | 6 days | CGM, intravenous BG data were collected five times a day | Free on request [41] | 3 min | [41] |
| N/A | D1NAMO [91] | 9/ 26-45 | 4 days | CGM, ACC, ECG, breathing, annotated food pictures | Free to download | 5 min | [29,92] |
| N/A | Data collected for METABO (M2) | 10/ (41.8±4.39) | 10 days | CGM, PA every 1 min, multiple-dose insulin therapy | Private | 5 min | [25] |
| N/A | Private by Oviedo et al. (PO) | 10/ (41±10) | 786 days | CGM, insulin dose, CHO intake | Private | N/A | [67,68] |
| NCT03711656 | Private by Bertachi et al. (PB) | 10/ (31.8 ±16.8) | 12 weeks | CGM, manual insulin dose, HR, step count, estimation of burned calories, estimated CHO intake | Private | N/A | [69,76] |
| NCT03643692 | ARISES (A) | 12/ 28-48 | 6 weeks | CGM, insulin dose, physiological signals | Private | 5min | [38] |
| N/A | ShanghaiT1D (ST1D) [93] | 12/ 37-68 | 3 - 14 days | CGM, insulin dose, CHO intake, demographic data | Free to download [94] | 5 min | [45,54] |
| N/A | Private by Ben Ali et al. (PBA) | 12/ N/A | 14 days | CGM | Private | 15 min | [26] |
| NCT03643692 | Private by Zhu et al. (PZ) [39] | 12/ 30–49 | 6 weeks | CGM, insulin dose, stress, illness, EDA, IBI, ACC, and skin temperature signals, BVP, HRV every 5 min, PA every 5 min, skin tempt 5 min, CHO intake | Free on request | N/A | [39] |
| N/A | Private Mazgouti et al. (PMZ) | 12/ N/A | 14 days | CGM | Private | 15 min | [65] |
| N/A | Yunnan (YT1D) | 13/ N/A | 4 weeks | Meal intake (meal times, meal components CHO), demographics, insulin, CGM | Private | 5 min | [64] |
| N/A | Private Giancotti et al. (PGT) | 17/ (12.3 ±4.3) | 200 days in total | CGM, HR | Private | 15 min | [59] |
| N/A | Private Marx et al. (PMX) | 17/ 7-16 | 6 days | CGM, demographic data, exercised in supervised sports camp; PA (type, duration), insulin dose, CHO intake, symptoms of hypoglycemia and self-monitoring BGL were noted in a logbook | Private | 5 or 15 min | [43,84] |
| NCT02862730 | Private by Tyler et al. (PT) [37] | 20/ (34.5±4.7) | 4 days | CGM, performed 8 identical exercises (sessions of ca. 40 min), ACC, HR, insulin dose | Free on request [37,95] | N/A | [37] |
| N/A | T1Dexi | 20/ (32±13) | 139 sessions | CGM, PA, insulin dose | Private | N/A | [81] |

*ACC, Accelerometer data. BG, Blood Glucose. BGL, Blood Glucose Levels. BVP, Blood Volume Pulse. CGM, Continuous Glucose Monitoring. CHO, Carbohydrate. CT, Clinical Trial. EDA, Electrodermal Activity Data. ECG, Electrocardiogram Data. GSR, Galvanic Skin Response Data. HR, Heartrate. HRV, Heartrate Variation. IBI, Inter-Beat-Interval. MACC, Magnitude of Accelerometer. N/A, Not Available. PA, Physical Activity. PN, Patient Number. STD, Standard Variation.*

The medium-sized datasets range from 25 to 70 subjects and comprise 11 datasets in total, of which 6 are openly accessible. Similar to the small-sized datasets, data involving children and adolescents are limited, with only three studies focusing on these age groups. Although study durations tend to be shorter compared to those in the small volume datasets, multiple variables are typically collected, often including data of physical activity, CHO, and insulin. Also, while most datasets are collected under free-living conditions, two are obtained in a hospital setting.

*Table 4: Medium Volume Datasets Used by Presented Studies for Type 1 Diabetes*

| **CT Number** | **Dataset** | **PN/ Age (Mean ± STD)** | **Study Duration** | **Description** | **Availability** | **CGM Frequency** | **Used by** |
|---|---|---|---|---|---|---|---|
| N/A | AI4PG [96] | 25/ (38±12) | 6-7 days | CGM, dietary habit and meal intake components, insulin dose, food diaries, meals as time series of pre- and post-meal glycemic levels (mg/dL) | Free on request [96] | N/A | [40] |
| NCT02053051 | ABC4D | 25 / (40±12) | 6 months | CGM, PA, insulin dose, CHO intake | Private | 5 min | [38] |
| N/A | Data collected for METABO (M1) | 27/ 19-72 | 5 - 22 days | CGM, PA and physiological signals every 1 min, CHO intake, insulin dose, insulin type | Private | 5 min | [24] |
| N/A | GlucoseML (GML) | 29/ 19-61 | 4 weeks | CGM | Private | 1 min | [34] |
| N/A | Open APS Data Commons | ≥19/ 11-60 | 96-1688 days per subject | CGM, insulin, CHO intake | Free on request [50] | N/A | [50] |
| N/A | DirectNet (DN07) [98] | 30-50/ 3-7 or 12-18 | 1 week | CGM | Free to download | 5 min | [9,28, 40,44] |
| N/A | Tidepool (TP2) | 50/ (38±13) | 21272 days in total (CGM+ insulin) | CGM, demographic data, insulin dose, self-reported PA (type, duration, timing, and intensity, ca. 170 days per subject) | Private | 5 min | [81] |
| N/A | DirecNet (DN05) [101] | 50/ (14.8±1.7) | 1 day | Hospital stay: CGM, 5 to 15 min walk in the morning, 75-minute exercise session in the late afternoon, insulin dose (27 subjects) | Free on request [101] | 5 min | [7] |
| N/A | Diatrend (DT) [102] | 54/ 19–74 | 27 days | CGM | Free to download [103] | N/A | [58] |
| N/A | DIAdvisor (DA) | 59/ (43.4±11.7) | 3 days | Hospital stay: glucose, insulin dose and correction doses, meal nutrients content for CHO, protein, and lipids | Private | N/A | [33] |
| N/A | UCI Machine Learning Repository, 2005 (UCI2005) [104] | 70/ N/A | N/A | CGM, activity during glucose measurement. number of NPH insulin shots per day, number of snacks | Free to download [104] | N/A | [77,79] |

*CGM, Continuous Glucose Monitoring. CHO, Carbohydrate. CT, Clinical Trial. BMI, Body-Mass-Index. HR, Heartrate. N/A, Not Available. PA, Physical Activity. PN, Patient Number. STD, Standard Deviation.*

The large datasets, ranging from 74 to 9800 subjects, comprise a total of 15 datasets. Of these, 10 are available either upon request or as free downloads. In this group, sensor data collection is typically limited to CGM, and study durations are generally longer than those in the smaller dataset cohorts. As the cohort size increases, the number of recorded features often decreases. Notably, five studies in this group include data from children and teenagers. Most datasets in this category are collected under free-living conditions, are voluntarily uploaded by participants. Only one study requires participants to engage in structured exercise sessions.

Finally, considering all datasets, where reported, the frequency of CGM measurements ranges from 1 to 15 minutes, with a 5-min sampling interval being the most commonly used. Only one study leverages sensors with a 1-min sampling rate. For other wearable data, such as HR, a 1-minute sampling interval is most frequently applied.

*Table 5: Large Volume Datasets Used by Presented Studies for Type 1 Diabetes*

| CT Number | Dataset | PN/ Age (Mean ± STD) | Study Duration | Description | Availability | CGM Frequency | Used by |
|---|---|---|---|---|---|---|---|
| NCT03844789 | Private by Breton et al. (PBre) [105] | 101/ 8-13 | 16 weeks | CGM, insulin (78) | Free on request [105] | 5 min | [53] |
| NCT01514292 | Private by Christiansen et al. (PC) [106] | 103/ 18-74 | 7 days | CGM | Private | 5 min | [107] |
| N/A | Private by Dave et al. (PD) | 112/ 1-21 | 90 days | CGM, insulin dose and estimated CHO intake (19 subjects) | Private | N/A | [70] |
| NCT03263494 | CITY [108] | 153/ 14-24 | 26 weeks | CGM, finger prick glucose | Free to download [109] | 5 min | [83] |
| NCT03563313 | Private by Brown et al. (PBro) [110] | 118/ 14-71 | 26 weeks | CGM, insulin (102) | Free on request [110] | 5 min | [53] |
| N/A | Tidepool (TP1) | 120/ N/A | 3 years | CGM | Private | 5 min | [82] |
| N/A | Private dataset from Samsung medical center (SMC) [31] | 141/ (55.0±13.4) | 3 days | CGM, demographic data (age, sex, BMI), duration of diabetes, medication | Available from authors upon request and with permission of SMC [31] | N/A | [31] |
| N/A | OpenAPS by Shena and Kleinberg (OAPSSK) | 183/ N/A | N/A | CGM, insulin, CHO | N/A | 5 min | [61] |
| NCT03240432 | WISDM [111] | 206/ 60-83 | 6 months | CGM | Free to download [112] | 5 min | [83] |
| NCT02258373 | ReplaceBG (RBG) [113] | 226/ 19-78 | 26 weeks | glucose from CGM (149 subjects), glucose from CGM and finger picks (77 subjects), insulin dose, CHO intake | Free on request [114] | 5 min | [33,82, 83] |
| 2023623235 | RICEL | 406/ 27-50 | 6-7 days | CGM | Free on request [80] | N/A | [80,11 5] |
| NCT00406133 | Juvenile Diabetes Research Foundation (JDRF) [116] | 451/ > 9 | 12 months | CGM | Free on request | 5 min (355) and 10 min (96) | [53,82, 98] |
| NCT02825251 | Onset 5 trail by Novo Nordisk A/S (NN) [117] | 472/ (43±15) | 16 weeks | CGM, insulin dose, CHO intake, demographic data | Free on request | N/A | [71,11 7,118] |
| N/A | T1Dexi by Neumann et al. (T1DEXIN) | 497/ ≥ 18 | ≥ 2 years | At least six structured exercise sessions over four weeks, (socio)demographic data, CGM, insulin, CHO, HR, PA | Private but available on request by Vivli Inc [119] | 5 min | [62] |
| N/A | Private by Vu et al. (PV) | 9800/ (45.3±16.4) | 100 days | voluntarily uploaded CGM data, insulin dose, CHO intake | Private | N/A | [72] |

*CGM, Continuous Glucose Monitoring. CHO, Carbohydrate. CT, Clinical Trial. N/A, Not Available. NPH, Neutral Protamine Hagedorn Insulin. PA, Physical Activity. PN, Patient Number. STD, Standard Deviation*

# IV. Discussion and Limitations

This study presented the current research state of glucose forecasting and hypoglycemia classification for short- and long-term PHs. Coming back to the research questions, it could be identified that: 1) In both use cases, a PH of up to 1 hour yields the best performance. 2) The best architectures for regression were achieved with stacked models, particularly Bi and Kar obtained the best 1 h prediction performance using both ML and DL as base-models [9,49]. FFNN obtained good results up to 45 min, whereas LSTM networks were best for Nasir et al. and Zafar et al. for 15 min and 48 h, respectively [30,40,44,50]. It was reported that the model's performance depends on the PH [43]. Thus, most studies are not robust across multiple PHs with one model. Stable results are only seen in Hakim et al., Georga et al., Xiong et al., and Mazgouti et al., approaching ML methods [24,54], combining transformers with a BiLSTM model [64], or leveraging an ensemble model of LSTM and XGBoost [65]. Allam showed that WOA is better for forecasting glucose values up to 45 min, while an RNN is superior for PHs larger than 45 min [30,51]. Furthermore, DL can tend to overfit and work better if trained on larger datasets.
For classification, conventional ML methods performed better with SVMs and RFs being most popular. A stacked model was only tested in D'Antoni et al. for short-term PHs, not presenting good performance, which could be impacted by the dataset's imbalance. In general, 70-80% of hypoglycemic events could be classified in all PHs. However, instead of using one model for multiple PHs, usually different datasets and models for different use cases were considered. No model was tested across short- and long-term PHs. Only Dave et al. classified from 15-60 min with the same model and obtained robust performance. For the PH, they considered horizon ranges instead of exact times. 3) It could be identified that the size, features, and quality of the dataset impact the performance, especially for longer PHs starting from 30 min. Impactful features are insulin data, meal information, laboratory data, demographics, and wearable signals. Notably, activity data is effective for predicting hypoglycemic ranges. Particularly, the timing and intensity of PA could influence the classification performance for long-term PHs (6 h). Nighttime models yield better results, and it could be observed that each use case may require a different set of features. 6 out of the analyzed 56 studies use the output of mathematical insulin, carbohydrate, or activity models as input for ML [24,68–70,76,77] and two for DL [29,38]. Only one study concatenated static data with DL. Since their dataset seemed to produce worse results, their model is suggested to be tested or validated on other datasets [43]. Looking at the ISL, most often the same length as the PH is utilized, and short-term prediction tends to use a length of 30 min. One could try a history window twice as long as the PH for short-term and as long as the PH for long-term classification. Nevertheless, an optimal way for determining the ISL cannot be identified. Only single studies explore different lengths per PH or subject, which is encouraged. Piersanti et al., achieving the best performance for a 24 h exercise risk assessment, use less information from 2-5 h of prior data [7], indicating that the ISL should be examined depending on the model, data, and use case. Longer input sequences could provide more helpful information, while too long sequences can decrease the performance. 4) Static data, including demographics and patient profiles, have significant impact on the performance. Moreover, different model architectures, feature combinations, and ISL perform differently for each subject, and this is why personalization can lead to better performance. It was shown that individual input selection improved performance and that different input sets are presented for different clusters of patients or different individuals [40,57].

Finally, some research gaps are identified. The primary challenge is the limited size of multivariate datasets, which often causes overfitting in DL models, particularly with datasets like OhioT1DM. Ghimire et al. showed that generalization is possible with larger datasets [53]. Currently, available multivariate datasets of adults have mostly data collected from 10-15 participants. While larger datasets exist, they frequently lack feature diversity and include more missing values. Vital signs such as ECG, HR, GSR, and IBI are rarely collected, despite studies reporting correlations with hypoglycemia and HRV, and glucose and ECG [91,120]. It was already identified that the feature set has a significant impact on the performance of the selected use case. Combining data from multiple sensors such as ACC, BVP, GSR, ST, and HR enhances classification stability over single-sensor approaches [121]. In particular, nocturnal and exercise-based hypoglycemia are improved by physiological data. For exercise-related hypoglycemia, activity data is significant. The type and intensity of PA and the inclusion of stress data improve the 1 h forecasting performance [20]. Nevertheless, behavioral, psychological, and activity-related parameters remain underexplored. Only Tyler et al. classified into lower and higher aerobic fitness [37].
Second, larger datasets can increase intrapersonal variability, whereas insufficient individual data reduces the performance of personalized models since interpersonal variations are less represented. Notably, studies focusing on hypoglycemia face this shortcoming and leverage synthetic data, which overestimate the results [29,56]. Thus, population-based models likely perform better. Personal features are stressed to have an impact, but most of the featured studies classifying hypoglycemia lack clustering or preprocessing methods. Clustering based on common comorbidities, demographics, or glucose trends to represent similar or individual profiles can enhance population-based models. Yang et al. demonstrated that age-based clustering is beneficial [57], while also clustering based on clinical characteristics can improve individualization with limited data [122]. Additionally, no studies combine datasets to capture the heterogeneous behaviors of a broader population. Only a few studies validate their models with external datasets. It is asserted that the model's performance is highly dependent on the dataset, highlighting the dependence of performance on dataset quality. The third limitation is the possible time lag of CGM devices between 5-12 min [26,123]. Hence, regression models require high accuracy, but presented studies report mean RMSEs of 10–15 mg/dL for 30 min predictions. Accurate results lack further validation on external datasets. In general, a 120 min prediction presents insufficient performance for clinical use cases. Contrariwise, the forecasting of 15 min is accurate but could be insufficient to prevent hypoglycemia on time. Classification models, by contrast, analyze patterns in data and focus on identifying adverse event risks. As to why, they are less affected by numerical differences, mitigating wearable data limitations [71]. They present good recall, but the precision and false alarm rates require improvement, possibly due to the underrepresentation of hypoglycemia. Notably, regression models are more prevalent than classification models, highlighting a research gap.

Coming to the model architectures and improvement strategies, featured works prefer DL for regression and ML for classification. For regression, 89% of studies approach NN, DL, or EL, whereas the performance is not consistent across PHs. Bi and Kar highlight that both approaches have advantages and disadvantages, suggesting their potential as ensemble models for improved performance. For classification, EL and DL are employed by only 38% of the studies. Poorer results on larger datasets may be attributed to the limitations of ML models, as DL can offer more effective analysis for big data. Although 48% of studies include larger cohorts of at least 70 subjects, only two studies utilize EL for short-term PHs [75]. More complex algorithms, such as stacked models, are only considered by D'Antoni et al. for short-term PHs, and hybrid models are utilized by Felizardo et al., yielding better results than using the models separately for long-term PHs. Expanding conventional models also results in some improvements, but is only applied by Mosquera Lopez et al. [81]. In addition, only one study generates time series data into images for image classification [78]. Consequently, it is encouraged to approach both ML and DL strategies and to investigate EL for long-term PHs and for classification. Lastly, explainability is rarely investigated and could be further studied for patient awareness.

Fourth, only a few studies leverage data from children. The onset of T1D is usually in childhood, and children are at increased risk for exercise-induced hypoglycemia [7,22]. Data from children may be more complex, as glucose levels and insulin needs can vary significantly across different age groups [22,43]. Moreover, only two studies trained models on both adult and children's data or explored the feasibility of transfer learning between these cohorts, but with insufficient performance or validation [41,53]. Also, details of the dataset are not always provided. Therefore, further research is encouraged to investigate transfer learning between age groups and determine whether children and adults require distinct models due to differences in behavior. Moreover, hypoglycemia is defined with a different threshold than the standard of 70 mg/dL in some studies due to variations in children's data or the time lag of CGM devices. However, no study was seen investigating individual thresholds, such as in Bent et al., for data of type 2 prediabetes patients [124].

The fifth observed shortcoming is that most studies focus on short-term PHs of 30-60 min, especially for regression. These can enable preventive self-actions, but cannot be used for insulin administration, suitable meal suggestions, and exercise duration predictions listed as the main causes of hypoglycemia. Classification models aim to classify hypoglycemia in a specific use case and thus, focus mainly on nocturnal hypoglycemia (6-9 h) or the risk assessment 24 hours prior. Most approaches are based on binary classification or at most three classes. None of the presented studies integrates multiple PHs into one model, which should be possible with a classification system such as in Cinar et al. and Onwuchekwa et al. [125,126]. Hence, the main motivation of developing a multifunctional classification system remains a research gap. The individual use cases of continuous monitoring, postprandial and nocturnal hypoglycemia, and the risk assessment of hypoglycemia for longer PHs could be classified up to 70-80% each.

Finally, the variety of metrics disables proper comparison. For regression, not all studies directly reported the metric of the RMSE value, which is essential for glucose forecasting. Moreover, 3 studies [30,43,51] reported RMSE values only in mmol/L, which had to be converted by the authors of this review with the formula: mmol/L * 18.018 = mg/dL [127]. As one of the standard metrics, it is suggested to report values in mg/dL and to define the metric in the report. The CEGA is an important metric as well, featured by most of the presented studies, but was out of the scope for this review. Likewise, a thorough comparison is disabled with classification models. The main metrics that should ideally be included are precision, recall, sensitivity, specificity, and F1-measure.

## V. Conclusion and Future Work

In summary, some studies can achieve clinically sufficient and accurate results, even for a 24-hour classification. However, these approaches have not been verified with additional datasets. The main limitation in hypoglycemia research is the underrepresentation of hypoglycemic datapoints, compared to euglycemia and hyperglycemia, which results in low precision. Another limitation is the size of multivariate datasets. The quality, size, and features of a dataset, as well as the age group of the population, can significantly impact performance, and the features need to be selected depending on the investigated use case. For regression, stacked models, ensemble models and meta-learning benefitting from multiple architectures and mitigating the disadvantages seem superior. The possibility of training one model for short- and long-term PHs is only investigated by 4 studies for regression, but not for classification. For classification, only single models for specific tasks are reported. One model explored a one-week-ahead prediction and showed that prior days of 0-3 days ago yield accurate predictions only. Hence, it is suggested to use individual models for short- and long-term, as well as for multiple long-term PHs. Furthermore, ensemble models, meta-learners, and stacked models should be developed for hypoglycemia classification with larger datasets and longer PHs. Personalized approaches, data preprocessing, and clustering were also not explored by most of the studies. Finally, the combination of datasets and transfer learning between different age groups and datasets is proposed.

In future work, we aim to address the limitations of non-generalized and insufficiently validated models by integrating accessible datasets containing CGM data from individuals with T1D. Publicly available datasets used in prior studies, including those referenced in this work, will be identified, processed, and integrated into a comprehensive CGM dataset. This dataset should ideally include a large and diverse population, spanning various age groups and genders, to improve generalizability. Using this expanded dataset, an exploratory comparison of state-of-the-art ML and DL models for both regression and classification tasks can be performed. Additionally, individualized models may be developed for individuals with a higher volume of available data. Subject-specific analyses, such as demographic clustering based on CGM-derived features, can also be conducted to assess whether age-dependent or demographic-specific models offer improved performance, or whether a robust, population-independent model can be achieved.